\let\ifarxiv\iffalse
\let\ifanon\iffalse
\let\ifdraft\iffalse
\let\ifcameraready\iftrue

\documentclass{article} %
\usepackage{iclr2026_conference}
\ifcameraready\iclrfinalcopy\fi

\usepackage{times}

\usepackage[hyphens]{url}  %
\usepackage{graphicx} %

\usepackage{caption} %

\usepackage[colorlinks=true, linkcolor=blue, citecolor=blue, urlcolor=blue]{hyperref}

\usepackage{algorithm}
\usepackage{algorithmic}

\usepackage{wrapfig}
\usepackage{newfloat}
\usepackage{listings}
\DeclareCaptionStyle{ruled}{labelfont=normalfont,labelsep=colon,strut=off} %
\floatstyle{ruled}
\newfloat{listing}{tb}{lst}{}
\floatname{listing}{Listing}

\usepackage{amsmath,amssymb}
\usepackage{framed}

\usepackage{tabularray}
\UseTblrLibrary{booktabs}
\SetTblrInner[booktabs]{abovesep=0pt, belowsep=0pt, rowsep=0.2pt}
\NewTableCommand\seprule{\specialrule{\lightrulewidth,gray8}{2pt}{2pt}}
\NewTableCommand\uniquerule{\specialrule{\lightrulewidth,gray7,dashed}{2pt}{2pt}}

\usepackage{subcaption}
\usepackage{enumitem}
\usepackage{tikz}
\usetikzlibrary{positioning,arrows.meta,calc}

\usepackage{cleveref}
\Crefname{section}{\S}{\S\S}
\crefformat{section}{\S#2#1#3}
\Crefname{appendix}{\S}{\S\S}
\crefformat{appendix}{\S#2#1#3}
\Crefname{figure}{Figure}{Figures}
\crefformat{figure}{Figure #2#1#3}
\Crefname{Figure}{Figure}{Figures}
\Crefname{Table}{Table}{Tables}
\crefformat{table}{Table #2#1#3}

\lstdefinestyle{codeblock}{
    basicstyle=\ttfamily\footnotesize,
    backgroundcolor=\color{white},
    commentstyle=\color{red!60!black},
    keywordstyle=\color{green!50!black},
    stringstyle=\color{red!60!black},
    basicstyle=\ttfamily\footnotesize,
    breakatwhitespace=false,
    breaklines=true,
    captionpos=b,
    keepspaces=true,
    showspaces=false,
    showstringspaces=false,
    showtabs=false,
    tabsize=2
}

\title{
Agnostics:
Learning to Synthesize Code in Any Programming Language with a Universal Reinforcement Learning Environment
}
\author{
    Aleksander Boruch-Gruszecki,\!\textsuperscript{\rm 1*}
    Yangtian Zi,\!\textsuperscript{\rm 1}
    Zixuan Wu,\!\textsuperscript{\rm 1}
    Tejas Oberoi,\!\textsuperscript{\rm 2}\\
    \textbf{
    Carolyn Jane Anderson,\!\textsuperscript{\rm 3}
    Joydeep Biswas,\!\textsuperscript{\rm 2}
    Arjun Guha\textsuperscript{\rm 1}
    }\\
    \textsuperscript{\rm 1}Northeastern University,
    \textsuperscript{\rm 2}University of Texas,
    \textsuperscript{\rm 3}Wellesley College\\
    \textsuperscript{\rm *}research@abgru.me
}

\newcommand\OurTechnique{Agnostics}
\newcommand\OurDsMbpp{Ag-MBPP-X}
\newcommand\OurDsCodeforces{Ag-Codeforces-X}
\newcommand\OurDsCodeforcesS[1]{Ag-Codeforces-#1}
\newcommand\OurDsLcb{Ag-LiveCodeBench-X}
\newcommand\OurDsLcbS[1]{Ag-LiveCodeBench-#1}

\begin{document}

\maketitle

\begin{abstract}
Large language models (LLMs) excel at writing code in \emph{high-resource} languages such as Python and JavaScript, yet stumble on \emph{low-resource} languages that remain essential to science and engineering. Besides the obvious shortage of pre-training data, post-training itself is a bottleneck: every new language seems to require new datasets, test harnesses, and reinforcement-learning (RL) infrastructure.

We introduce \OurTechnique, a \emph{language-agnostic} post-training pipeline that eliminates this per-language engineering. The key idea is to judge code solely by its externally observable behavior, so a single verifier can test solutions written in \emph{any} language. Concretely, we
(i) use an LLM to rewrite existing unit-test datasets into an I/O format,
(ii) supply a short configuration that tells the verifier how to compile and run a target language, and
(iii)~apply reinforcement learning with verifiable rewards (RLVR) in a robust code execution environment.

Applied to five low-resource languages—Lua, Julia, R, OCaml, and Fortran—\OurTechnique{}
(1) improves Qwen-3 4B to performance rivaling other 16B–70B open-weight models;
(2) scales to larger and diverse model families (Qwen-3 8B, DeepSeek Coder 6.7B Instruct, SmolLM 3, Phi 4 Mini); and
(3) for ${\le} 16$B parameter models, sets new state-of-the-art pass@1 results on MultiPL-E and a new multi-language version of LiveCodeBench which we introduce.

We release the language-agnostic training datasets (\OurDsMbpp, \OurDsCodeforces, \OurDsLcb), training code, and ready-to-use configurations,
making RL training for \emph{any} programming language as easy as a few lines of YAML.
\end{abstract}

\section{Introduction}

Large language models (LLMs) are remarkably good at programming tasks,
especially when coding in \emph{high-resource programming languages} such as Python and JavaScript.
Their proficiency in \emph{low-resource programming languages}, such as Fortran, Julia, and others,
is far more limited.
This gap appears both on benchmarks~\citep{cassano:multipl-e} and in popular discourse.
Many low-resource languages are adapted to and widely used in particular sectors
such as computational science (e.g., Julia, Fortran), medicine (e.g., Mumps), data science (e.g., R), and others.
Methods for improving LLMs on such languages
would help programmers in these sectors truly take advantage of LLMs.

The capability gap between high-resource and low-resource programming languages occurs for two reasons.
First, there is \emph{vastly} more training data for some languages.
For example, The Stack V2~\citep{lozhkov2024}, the largest public training corpus of code,
has ${\approx} 200$GB of Python but only ${\approx} 2$GB of Julia and Fortran.
Thus pretraining on code makes models significantly better at Python.
A subtler reason is the availability of post-training datasets and techniques.
Contemporary LLMs are developed with an extensive post-training process that relies on
(a)~high-quality curated data for supervised fine-tuning,
and (b)~carefully designed environments for reinforcement learning,
which must be able to execute and verify model-generated solutions.
Both of these require significant human expertise,
which is hard to find for low-resource programming languages.

Our goal in this work is to facilitate post-training LLMs on low-resource programming languages,
working towards closing the resource gap.
Our key idea is that for a large class of programming tasks,
correctness can be stated as a property not of functions or code snippets,
but of the entire program's observable behavior (e.g., I/O).
Furthermore, if its correctness can be tested with a \emph{verifier} program,
such a \emph{verifier} with appropriate problems and test cases
can be used to make a universal reinforcement learning environment
which can be instantiated for nearly any programming language.
In fact, the verifier's implementation language
is independent from the one being learned.
This approach matches the formulation of some existing post-training datasets
(even if they are intended for Python/C++),
and we can reformulate language-specific datasets into this format with LLMs.

\begin{figure*}[htbp]
\includegraphics[width=\textwidth]{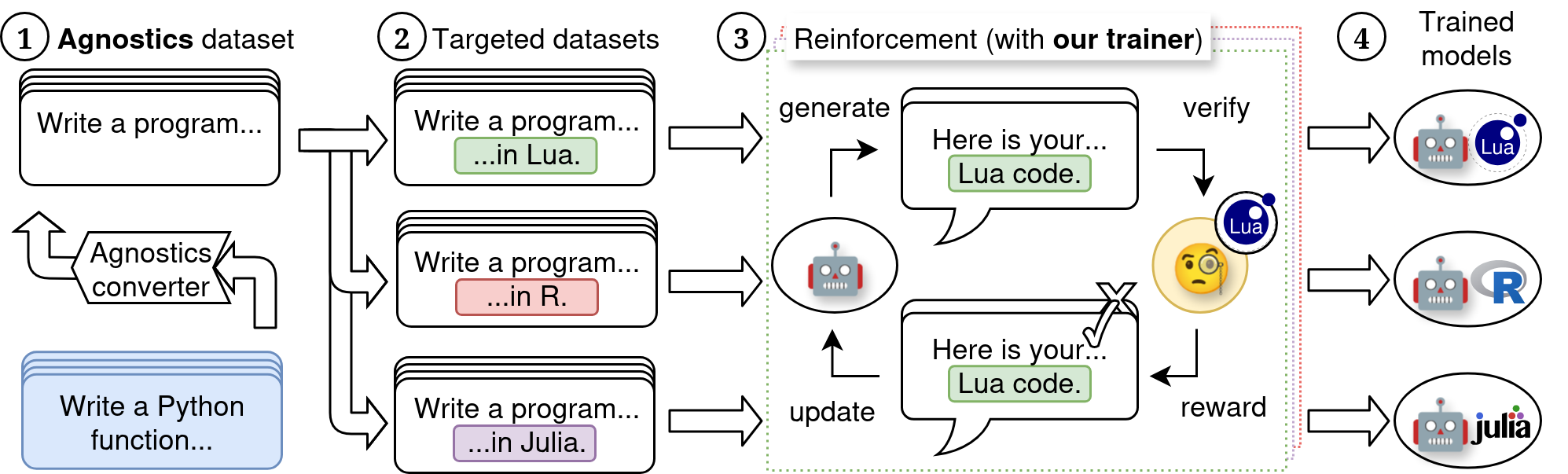}
\caption{
  \textbf{Overview: Agnostics Data Preparation and Training.}
  \textbf{(1)}~We reformulate existing coding datasets to our format.
  \textbf{(2)}~We adapt the language-agnostic datasets to a particular programming language.
  \textbf{(3, 4)}~We reinforce coding via Group Relative Policy Optimization \citep{shao2024,deepseek-ai2025}, verifying the programs in our code execution sandbox.
}
\label{fig:overview}
\end{figure*}

Our approach, \OurTechnique{}, works based on this insight as follows (\Cref{fig:overview}).
1)~We use an LLM to reformulate language-specific datasets into our uniform language-agnostic format.
2)~To target a particular language, we generate prompts and instantiate the verifier
based on a small (4-5 line) configuration file.
3)~We apply reinforcement learning with verified rewards (RLVR) using a robust,
language-agnostic execution sandbox that we develop.
4)~The result is a model specialized to the target language.
\OurTechnique{} particularly excels at finetuning models for low-resource languages,
as it does not rely on high-quality datasets specific to a particular language.

\paragraph{Contributions}
\label{sec:contributions}
\begin{enumerate}[leftmargin=*, itemsep=0.5ex, topsep=0.5ex]
    \item \OurTechnique, a post-training pipeline for coding in arbitrary programming languages;%
    \item The best-performing open-weights ${\le} 16$B models for Lua, R, Julia, OCaml and Fortran;
    \item Three \OurTechnique{} datasets: \OurDsMbpp, \OurDsCodeforces, and \OurDsLcb,
    based on MBPP~\citep{austin2021}, Open-R1 Codeforces~\citep{penedo2025codeforces}
    and LiveCodeBench~\citep{jain:livecodebench} respectively.
    \item A small and carefully designed \OurTechnique{} training framework,
    including a parallel code execution sandbox, sampling,
    rewards computation, GRPO, and model back-propagation.
\end{enumerate}

\ifcameraready
Our leaderboard, data, code, and artifacts
are referenced at \href{https://agnostics.abgru.me}{agnostics.abgru.me}
and in \cref{sec:reproducibility}.
\fi

\ifarxiv
Artifacts and code necessary to verify the results we present
are listed at \href{https://agnostics.abgru.me}{agnostics.abgru.me}.
The training framework will be publicly released together with
the formal publication of this work.
\fi

\section{Background and Related Work}

\paragraph{Data Scarcity for Low-Resource Languages}

\begin{wrapfigure}{r}{.47\textwidth}
\vspace{-1\intextsep}
\centering
\(
\begin{array}{|l|r||l|r|}
\hline
\textbf{Language} & \textbf{\%} & \textbf{Language} & \textbf{\%} \\
\hline
\text{JavaScript} & 17.04\,\% & \text{Lua}        &  0.53\,\% \\
\text{Java}       &  8.38\,\% & \text{R}          &  0.35\,\% \\
\text{C++}        &  5.41\,\% & \text{Fortran}    &  0.07\,\% \\
\text{Python}     &  3.56\,\% & \text{Julia}      &  0.10\,\% \\
\hline
\end{array}
\)
\caption{High-resource (left) and low-resource (right) languages in the Stack v2.}
\label{fig:stack-volume}
\vspace{-1.2\intextsep}
\end{wrapfigure}

General-purpose LLMs have been pretrained on code for several years,
both because LLMs are widely used for practical programming tasks,
and because pretraining on code improves their general reasoning abilities~\citep{yingwei:pretraining-on-code}.
Code-specialized models may be trained from a general-purpose model checkpoint 
or exclusively on code (e.g., \citet{polycoder,allal:santacoder}).

However, the publicly available pretraining data for code is
heavily skewed toward a handful of programming languages.
E.g., consider The Stack V2~\citep{lozhkov:starcoder2},
the largest public code pretraining dataset,
with code from GitHub and dozens of other sources.
The Stack V2 is dominated by relatively few programming languages:
just 10 of 619 languages account for over 90\% of the dataset.
We want to develop models for low-resource programming languages
that each account for ${\leq} 0.5\%$ of publicly available code (\cref{fig:stack-volume}).
We can imagine working around this data scarcity in a few ways. However, previous work shows that up-sampling low-resource languages during pretraining only leads to small benchmark improvements~\citep{orlanski2023pre},
while fine-tuning on the pre-training data for low-resource languages has negligible impact~\citep{cassano2024}.
Thus, it is not clear how to make further gains from existing natural data.

\paragraph{Synthetic Data for Low-Resource Languages}

If natural data is not available for a task, it is possible to use LLMs to generate synthetic fine-tuning data~\citep{wang-etal-2023-self-instruct}, and there are many techniques for building code-centric supervised fine-tuning datasets (e.g.~\citet{ziyang:wizard-coder,wei:magicoder,wei:starcoder2-self-instruct}) that work remarkably for Python.
Although these approaches could in principle be applied to any programming language, \citet{cassano2024} show that without distillation or verification (e.g., see \citet{hu2025pre,wei2024}), synthetic tasks and code for low-resource programming languages are low-quality, and models fine-tuned on them perform poorly.

MultiPL-T~\citep{cassano2024}, similar to TransCoder-ST~\citep{roziere:transcoder-st} and CMTrans~\citep{cmtrans}, couples synthetic data generation with verification using rejection sampling: it generates up $n$ candidate programs in a target, low-resource language and only fine-tune models on generations that pass hidden unit tests that it translates from Python. However, the MultiPL-T approach has two significant limitations. (1)~For the verifier to not reject all samples, the model must be able to generate a working program within $n$ attempts.
In MultiPL-T, ${\approx} 30\%$ of prompts produce a working program for $n \in \{50, 100\}$ attempts,
and the rest are discarded.
We train on much harder problems,
and estimate that rejection sampling would require
an order of magnitude more resources for a comparable acceptance rate
(\cref{sec:rejection-sampling-efficiency}).
(2)~For each low-resource language of interest,
MultiPL-T requires writing a little compiler
to translate test cases and function signatures from Python to the target language.
MultiPL-T only supports a limited set of built-in Python types (e.g., no classes)
and dictates that all Python types and values must faithfully map to the target language.
However, depending on the problem and language,
the natural data representation may not map cleanly to Python.
This can lead to peculiar, unidiomatic translations that require deep language expertise to get right.
The \OurTechnique\ approach is far easier to use than MultiPL-T,
and only requires the user to know how to compile and run a program in the target language from the shell.

\paragraph{Reinforcement Learning on Coding Tasks}

DeepSeek R1~\citep{deepseek-ai2025} popularized RL on LLMs with rule-based rewards,
instead of learned reward models.
R1 reports applying RL to coding tasks without further dataset details.
A number of papers apply RL to the NL to code task~\citep{zeng:acecoder,gehring:rlef,jain:mu-code}.
These techniques target Python
and show that RL can improve LLM capabilities beyond
what supervised fine-tuning allows alone.

However, the key benefit of RL is that it can train a model to do tasks for which high-quality supervised fine-tuning data is unavailable.
There are recent examples of using RL for code optimization~\citep{du:afterburner,nichols:rlpf},
resolve GitHub, issue resolution~\citep{wei:swerl},
and iterative development~\citep{zhou:sweet-rl}.
These papers target tasks in high-resource languages (C++, Java, and Python) whereas \OurTechnique{} targets several low-resource languages.

\section{The \OurTechnique{} Approach}

Our approach comprises
(1)~a \emph{data preparation} stage which
reformulates language-specific programming tasks to be language-agnostic,
and retargets language-agnostic datasets to a programming language of interest~(1, 2 in \cref{fig:overview});
and (2)~the \emph{training} stage which
uses the GRPO algorithm and an efficient, language-agnostic verification framework~(3, 4 in \cref{fig:overview}).
Our tasks ask for programs by describing their behavior.
The test cases are samples of this behavior,
and a verifier program can check if a solution behaves according to the sample.
In this paper, we limited ourselves to working with tasks asking for programs
which read data from the standard input,
compute a unique answer,
and write it to the standard output.
Hence, the datasets we prepared share one verifier.

\begin{figure}

\begin{subfigure}{0.5\textwidth}
\begin{lstlisting}[style=codeblock,language=Python]
# Write a python function to
# identify non-prime numbers.
def is_not_prime(n):
    ...


~\colorbox{lightgray}{assert is\_not\_prime(2) == False}~
~\colorbox{lightgray}{assert is\_not\_prime(10) == True}~
\end{lstlisting}
\vskip 1em

\caption{An MBPP task prompt and associated tests (in gray).}
\label{fig:mbpp-example}
\end{subfigure}%
\begin{subfigure}{0.5\textwidth}
\textbf{Instruction:} Given an integer $N$ ($N\ge2$), determine if it is a non-prime number.
Output `True' if the number is non-prime, `False' otherwise.
Input format: a single integer $N$ ($N \ge 2$).
Output format: a single line containing `True' or `False'.
\vskip 1em
\begin{center}
\(
\footnotesize
\begin{array}{|l|l|}
\hline
\textbf{Input} & \textbf{Output} \\
\hline
2 & \texttt{False} \\
10 & \texttt{True} \\
\hline
\end{array}
\)
 \end{center}

\caption{The task and tests reformulated for \OurTechnique.}
\label{fig:mbpp-reformulated}
\end{subfigure}

\caption{For dataset preparation, we use an LLM to reformulate fine-tuning datasets with language-specific prompts and tests (above) into equivalent language-agnostic programming tasks.}
\label{fig:mbpp-transformation-example}
\end{figure}

\subsection{Dataset Preparation}
\label{sec:preparation}

Some datasets, like Open-R1 Codeforces~\citep{penedo2025codeforces}, already define tasks in the desired I/O style.
More commonly, however, code datasets provide a set of unit tests.
\Cref{fig:mbpp-example} shows a representative item from MBPP:
it has a natural language problem description and a Python function signature that comprise the prompt,
and a suite of tests used to test model-generated code.
These datasets can be easily translated into the I/O format.

To make such problems language-agnostic and compatible with our verifier,
we prompt an LLM to reformulate each task so that the program communicates exclusively via plain‑text standard in and standard out.
We ask the model to spell out concrete I/O conventions---number of decimal places,
newline versus comma separators, ordering of values, and so on---so that the expected behavior is unambiguous.
\Cref{fig:mbpp-reformulated} shows the reformulated example.
\Cref{appendix:reformulation} presents the prompt we use to reformulate MBPP;
other datasets might require small changes to the prompt.

\subsection{Programming Language Preparation}

To prepare a new language, we author a small configuration file with two purposes.
First, it defines a \emph{prompt prefix} (prepended to each problem by the trainer)
which instructs the model to produce code in the target language.
Second, the configuration file specifies
the shell commands to install the language toolchain and run code.
In our experience,
a prompt prefix simply asking for a solution in language $L$ is enough
for more widespread languages with ${\ge}5\%$ base accuracy.
However, when starting from near-zero accuracy,
a longer prefix can help prevent common mistakes.
E.g., our R language configuration (\cref{fig:r-config}) features a longer prompt
explaining the quirks of I/O APIs in R.\!\!\footnote{
  There are 3 ways to run R, 3 I/O APIs, and only one portable way to read from standard in.
}

If a model barely knows a programming language, a good prefix can help it.
Still, writing the prefix takes manual effort.
For OCaml and Fortran, we let a base model generate several faulty snippets,
and asked a capable LLM (OpenAI o3) for advice based on the snippets with the following prompt.
\begin{quote}
\itshape What follows are several Fortran programs. You'll see that most of them are wrong.  Read them carefully and identify the Fortran programming mistakes that I'm making. Ignore algorithmic mistakes, and focus on my misconceptions about Fortran. Come up with advice on how I should program Fortran correctly. Distill this advice into 10-20 sentences.
\end{quote}
We use the resulting instructions verbatim (see \cref{appendix:configurations}) when training models.
The prefix only needed to slightly raise the model's train split performance;
base accuracy as low as $0.09\%$ was enough for the model to start learning (see \cref{sec:rejection-sampling-efficiency}).
Configuring the two languages took 1 hour each.

\begin{figure}
\begin{lstlisting}[style=codeblock]
install: apt-get install -y r-cran-tidyverse
filename: snippet.R
execute: Rscript snippet.R
prompt: |
  Use R version 4. Use `readLines(con = file("stdin"))` to read from
  stdin. Use the `n` argument to read the first `n` lines. For example:
  ```r
  input <- readLines(con = file("stdin"), n = 1)
  n <- as.integer(input)
  cat(n) # print the first line of input
  ```
  Also, please remember to use `cat` to print output.
\end{lstlisting}
\caption{An \OurTechnique\ configuration snippet for R (slightly rephrased for presentation).}
\label{fig:r-config}
\end{figure}

\subsection{Trainer and Code Execution}
\label{sec:trainer}

The \OurTechnique{} trainer uses the Group-Relative Policy Optimization (GRPO) reinforcement learning algorithm~\citep{shao2024}, with verifiable rewards~\citep{deepseek-ai2025}, and further common tweaks to improve its efficiency~\citep{yu_dapo2025}. We couple the algorithm with a language-agnostic code execution framework designed to be robust and efficient.

\paragraph{Trainer}

The trainer instantiates the GRPO algorithm as follows.
Let $(x, \{(\mathit{in}_k,\mathit{out}_k)\}_{k=1}^K) \sim \mathcal{D}$ be a dataset of language-agnostics tasks,
where $x$ is the task prompt and $\{(\mathit{in}_k,\mathit{out}_k)\}_{k=1}^K$ is the set of I/O examples.
Let $P$ be $L.\texttt{prompt}$ from a language configuration $L$ (e.g., \cref{fig:r-config}).
From the behavior policy $\pi_{\theta_{\text{old}}}$ we sample a group $G$ of candidate responses
\(
\{y_i\}_{i=1}^{G} \sim
\pi_{\theta_{\text{old}}}({\cdot}\mid {P, x}).
\)
We assign each candidate a reward $R_i$,
with $R_i = 1$ if the execution environment (described later) verifies that the extracted program behaves as in
the I/O examples $(\mathit{in}_k,\mathit{out}_k)$ and $R_i=0$ otherwise.
We turn group rewards into sequence-level advantages $\hat{A}_i$,
and update the policy with the objective
\begingroup
\begin{align*}
\mathcal{L}_{\text{GRPO}}(\theta) &= \mathbb{E}_{\{
(x, \_) \sim \mathcal{D},
\{y_i\}_{i=1}^{G} \sim
\pi_{\theta_{\text{old}}}(\cdot \mid P, x)
\}} \\
&\quad \Biggl[
\frac{1}{G}\sum_{i=1}^{G}
\frac{1}{|y_i|}\sum_{t=1}^{|y_i|}
\min \Bigl(
\text{clip}\bigl(
r_{i,t}(\theta),1-\varepsilon,1+\varepsilon
\bigr)\hat{A}_i,
r_{i,t}(\theta)\hat{A}_i
\Bigr)
\Biggr],
\end{align*}
\endgroup
\[
  \text{where}
\qquad
r_{i,t}(\theta) = \frac{
  \pi_{\theta}(y_{i,t}\mid P,x,y_{i,<t})
}{
  \pi_{\theta_{\text{old}}}(y_{i,t} \mid P,x,y_{i,<t})
},
\qquad
\hat{A}_i = \frac{R_i - \operatorname{mean}(\{R_j\}_{j=1}^{G})}{\operatorname{std}(\{R_j\}_{j=1}^{G})}.
\]

We omit the KL-divergence term, similar to \citet{yu_dapo2025}.
We also considered and decided against a reward for a partially-correct answer (see \cref{appendix:reward-fn}).
We tried to reward the model
for code which runs without errors but produces wrong output
or for code which only passes the public tests (if there are any).
In both cases the models were very likely to learn how to exploit the reward,
e.g., by producing empty programs or by hard-coding the public tests
(and claiming to produce a ``draft answer'').

\paragraph{Code Execution}
Our verifier, a language-agnostic code execution sandbox,
(1)~extracts a program from each candidate;
(2)~compiles it if needed;
and (3)~tests it on the I/O examples $\{(\mathit{in}_k,\mathit{out}_k)\}_{k=1}^{K}$.

To extract the code, we instruct the model to put it in a Markdown block,
which all major instruction-tuned models do by default.
Since we rely on the native format of the model,
we do not need to train the model with a format reward.
This guarantees that the increases in rewards we see during training are real improvements
and not merely the result of the model learning to format correctly.

For each language, we build and cache an \citet{oci} container using the configuration $L$.
To build the container, we install the language compiler and runtime (the script $L.\texttt{install}$),
and include a generic execution harness which runs and tests candidate programs.
The execution harness runs continuously in the container,
waiting for triples with the candidate program,
the set of input/output examples, and timeouts.
The harness (1)~writes the program to disk (to $L.\texttt{filename}$),
(2)~compiles it if needed ($L.\texttt{compile}$),
(3) runs it on each received input ($L.\texttt{execute}$),
and verified that it produces the expected output.
The harness imposes timeouts on the compilation step and each execution,
and returns reward $0$ on any timeout or failed verification.
It is important to have timeouts for both compilation and execution. This prevents pathologies such as unbounded macro expansion in Julia
(caught by the compile timeout) and infinite loops (caught by
the execution timeout). Using containers also allows us to limit CPU, memory, and filesystem usage;
no elevated privileges are granted to the generated program.
Although the current datasets only specify tasks by standard I/O, the
same sandbox can safely accommodate problems that read/write from network or disk.

A subtle resource limit that we impose is a limit on the size of output.
Even with a short timeout such as $30$ seconds,
a pathological candidate program can output tens of gigabytes of text.
This can crash the verifier if it na\"{i}vely tries to read and store all output.
Instead, the verifier maintains a fixed-size (5MB) read buffer and
immediately kills programs which overflow it.

Overall, this design lets us keep a pool of warm containers for the duration of training:
we find that spawning a fresh container is two orders of magnitude slower than re-using an existing one.
In our experiments, a single training run may involve
testing 150,000 programs, each on several I/O examples.
Most of the generated programs are faulty and some behave badly,
e.g., they either timeout, consume too much memory, or produce too much output.
So containers do occasionally crash or need to be killed, and our execution environment handles this automatically.

Finally, to improve compile times,
our execution environment mounts a RAM disk in each container.
Compilation may be slow due to creating many intermediate files,
and indeed some large C++ projects, e.g., Firefox,
recommend using a RAM disk to speed up their builds~\citep{firefox-ramdisk}.

\paragraph{Implementation}
We implement the trainer and execution environment with Ray~\citep{moritz:ray}, which facilitates multiprocessing and distributed computing.
In particular, Ray lets us distribute the training
over a network of heterogeneous nodes,
which allows running the trainer on a node specialized for GPU work
and the execution environment on a node specialized for CPU work.
Ray also lets us easily separate group generation and loss computation
into inter-communicating processes.
Running the two in parallel significantly speeds up training,
as we found that they take a roughly comparable amount of time.
The execution environment is also an actor
and manages containers with Python \texttt{asyncio} coroutines, not actors, 
to minimize inter-process data copying.

\subsubsection{Hyperparameters}
\label{sec:hyperparameters}
We use the AdamW optimizer~\citep{loshchilov2019pre} with a learning rate of $5\times 10^{-6}$
and a cosine decay schedule with a warmup of 0.1 epochs.
We process 4 prompts in each batch, with group size 32 per prompt.
When training we use temperature 0.7
and disable reasoning in hybrid models.
(Many generations still show reasoning-like text, either before the answer or in comments.)
We describe how we chose the hyperparameter values and how we evaluted different values in \cref{appendix:choosing-hyperparameters}.

\section{Evaluation}
\label{sec:evaluation}

To evaluate \OurTechnique,
we train and benchmark models on 5 low-resource programming languages.
We measure pass@1 accuracy with reasoning disabled,
20 samples per prompt at temperature 0.2.
We trained the models for 1 epoch,
unless specified otherwise.

\subsection{Training datasets}

\textbf{\OurDsCodeforces}, the main dataset we use for training,
was created based on competitive programming problems from the Open-R1 Codeforces dataset~\citep{penedo2025codeforces}.
Few adjustments were necessary,
since the problems already specified programs and tests using standard I/O.
The train split contains 5369 problems.

\textbf{\OurDsMbpp}, the other training dataset we use, was created from MBPP as explained in~\cref{sec:preparation}.

Both datasets were analyzed to ensure no contamination with benchmarks we use.
See \cref{appendix:reformulation} for a more detailed description of how the datasets were prepared.

\subsection{Benchmarks}

We evaluate \OurTechnique{} with the following benchmarks.

\textbf{MultiPL-E}~\citep{cassano:multipl-e} is a well-established benchmark,
frequently used to evaluate the performance of new LLMs on a broad set of languages
(e.g.,~\citet{kimi-k2,yang:qwen3,llama3,yuyu:seedcoder}).
MultiPL-E was prepared by compiling HumanEval~\citep{chen:codex} prompts and unit tests from Python to each target language.
Each MultiPL-E programming language requires a ${\approx} 500$ LOC prompt and test translator,
considerably more effort than writing an \OurTechnique\ configuration file.
A major limitation of MultiPL-E is being too easy for frontier models.
With Python, frontier models are now evaluated on
solving programming contest problems~\citep{jain:livecodebench};
no multi-language benchmarks are as challenging.

\textbf{\OurDsLcb}, a contribution of this paper, is a new multi-language benchmark derived from LiveCodeBench.
(The benchmark has no overlap with our training data; see \cref{appendix:reformulation}.)
LiveCodeBench 5.0 has 880 problems;
381 problems have Python starter code and test cases,
while the remaining 499 problems instead use standard I/O to specify and test solutions.
We used the latter problems to transform LiveCodeBench into an \OurTechnique\ dataset.
Accordingly, benchmarking a new programming language with \OurDsLcb{} is straightforward: we can reuse our trainer's language configurations and execution environment (\cref{sec:trainer}).
Our results and leaderboard (see \cref{sec:contributions}) show that \OurDsLcb{} is far harder than MultiPL-E and is challenging for frontier models as of February 2026.

\subsection{Results}

\begin{wraptable}{r}{.36\textwidth}
\vspace{-1\intextsep}
\footnotesize %
\centering
\begin{booktabs}{colspec={@{}l|[gray8]rrr@{}}, colsep=3pt}
  \toprule
  Model & \SetCell[c=3]{c} Ag-LCB-X \\
  \SetCell{r} X= & Lua & Julia & R \\
  \midrule
  Llama 3.3 70B Ins
        & \textbf{25} & 22 & 13 \\
  Qwen 3 32B
        & 22 & \textbf{26} & \underline{17} \\
  DSC v2 Lite Ins 16B
        & 13 & 12 & 9 \\
  \seprule
  Qwen 3 4B
        & 11 & 10 & 10 \\
  Qwen 3 8B
        & 11 & 9 & 9 \\
  \SetRow{azure9}
  Qwen3-4B-MBPP-X
        & 15 & 15 & 9 \\
  \SetRow{azure9}
  Qwen3-4B-CF-X
        & \underline{23} & 22 & 15 \\
  \SetRow{azure9}
  Qwen3-8B-CF-X
        & \textbf{25} & \underline{25} & \textbf{19} \\
  \bottomrule
\end{booktabs}
\caption{
  Ag-LCB-X pass@1.
}
\label{tab:results-lcbx-main}
\vspace{-2.5\intextsep}
\end{wraptable}

We now present our results.
We use a few abbreviations in the tables.
Ag-LCB-X stands for \OurDsLcb;
we clarify abbreviated model names in the text.
Highlighted rows present our models;
note that each cell in such a row presents the score
of a \emph{different} model trained on programming language X.
We compute the score as explained in \cref{sec:evaluation}.

\paragraph{SOTA small LLMs for low-resource PLs}
Using \OurTechnique{}, we train Qwen 3 4B and 8B on \OurDsCodeforces\
specialized to Fortran, Julia, R, Lua, and OCaml.
The resulting models are
state-of-the-art 
at coding in low-resource programming languages
among
open-weight models with ${\le} 16$B parameters.

\begin{wraptable}{r}{.375\textwidth}
\vspace{-1\intextsep}
\small
\centering
\begin{booktabs}{colspec={@{}l|[gray8]rr@{}}, colsep=3pt}
  \toprule
  Model & \SetCell[c=2]{c} Ag-LCB-X \\
  \SetCell{r} X= & OCaml & Fortran \\
  \midrule
  \textit{Sonnet 4}
        & \underline{6} & 6 \\
  Llama 3.3 70B Ins
        & \textbf{7} & 3 \\
  Qwen 3 32B
        & 2 & 1 \\
  DSC v2 Lite Ins 16B
        & \textbf{7} & 6 \\
  \seprule
  Qwen 3 4B
        & 1 & 0 \\
  Qwen 3 8B
        & 0 & 0 \\
  \SetRow{azure9}
  Qwen3-4B-CF-X
        & \textbf{7} & \underline{15} \\
  \SetRow{azure9}
  Qwen3-8B-CF-X
        & \textbf{7} & \textbf{17} \\
  \bottomrule
\end{booktabs}
\caption{
  Ag-LCB-X pass@1.
}
\label{tab:results-lcbx-aux}
\vspace{-.75\intextsep}
\end{wraptable}

\OurTechnique\ training yields significant improvements 
on \OurDsLcb{} (\cref{tab:results-lcbx-main,tab:results-lcbx-aux}).
(i)~On every language, our models match or outperform DeepSeek Coder v2 Lite Instruct (16B),
and their performance comes close to or exceeds that of Qwen 3 32B and Llama 3.2 70B.
(ii)~The OCaml and Fortran scores improve from near zero to ${\approx}7\%$ and ${\approx}15\%$,
outperforming even some frontier models.
Importantly, these scores show what the models learned from training:
during evaluation
we omit the prompt prefix used to facilitate learning (\cref{sec:preparation}).
(iii)~Finally, the pass@1 scores improve by a factor of 1.5--2x over the base model.
This indicates that \OurTechnique\ can improve models beyond usual training on all the publicly available code data,
as we can safely assume that the Qwen models were trained on all such data, like the Llama models~\citep{llama3}.

\begin{wraptable}{r}{.355\textwidth}
\vspace{-1\intextsep}
\small
\centering
\begin{booktabs}{colspec={@{}l|[gray8]rrr@{}}, colsep=3pt}
  \toprule
  Model & \SetCell[c=3]{c} MultiPL-E \\
  \SetCell{r} X= & Lua & Julia & R \\
  \midrule
  Qwen 3 4B
        & 61 & 51 & 36 \\
  Qwen 3 8B
        & 63 & 53 & \underline{44} \\
  \SetRow{azure9}
  Qwen3-4B-MBPP-X
        & 51 & \textbf{62} & 41 \\
  \SetRow{azure9}
  Qwen3-4B-CF-X
        & \underline{64} & 54 & 43 \\
  \SetRow{azure9}
  Qwen3-8B-CF-X
        & \textbf{68} & \underline{61} & \textbf{52} \\
  \bottomrule
\end{booktabs}
\caption{
  MultiPL-E pass@1.
}
\label{tab:results-multiple}
\vspace{-1.5\intextsep}
\end{wraptable}

Models trained with our approach generalize over the competitive programming format:
the improvements are not limited to synthesizing programs using standard I/O.
To demonstrate this, we evaluate them on the established MultiPL-E benchmark.
It features problems which ask for Python functions operating on usual Python data structures,
and we find that our training also significantly improves the models on such problems
(\cref{tab:results-multiple}).\!\footnote{Note that MultiPL-E does not support OCaml and Fortran.}
We also confirmed that our training does not lower performance on other programming languages
(\cref{appendix:crosspl}).

\begin{wrapfigure}{R}{.5\textwidth}
\vspace{-1\intextsep}
\centering
\includegraphics[width=0.5\textwidth]{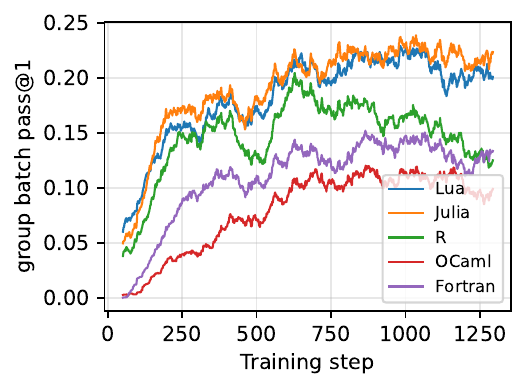}
\vspace{-2em}
\caption{Group batch pass@1, Qwen3-4B-CF-X.}
\label{fig:main-curves-all-train}
\vspace{-1\intextsep}
\end{wrapfigure}

\Cref{fig:main-curves-all-train} shows the GRPO batch pass@1 rates seen when training Qwen3-4B-CF-X.
All the models follow similar curves,
partially due to being trained on the same data permutation.
Nearly all the models slowly keep improving almost until the dataset end.
We also observed the train and test split rewards to be correlated with each other (\cref{appendix:training-curves}).

\paragraph{\OurTechnique{} scales to larger models}
To test if the gains from \OurTechnique\ training scale with model size,
we train the Qwen 3 8B model on \OurDsCodeforces\
specialized to Lua, Julia and R
and benchmarked it on \OurDsLcb\ and MultiPL-E
(\cref{tab:results-lcbx-main,tab:results-multiple}).
The Qwen3-8B-CF-X models show significant gains on both benchmarks,
improving over their 4B counterparts.
We expect \OurTechnique\ to scale to even larger models, with appropriate computing resources.
However, we found that \OurTechnique\ training on \OurDsCodeforces\ does \emph{not} improve two smaller models,
Qwen 3 1.7B and Llama 3.2 3B Instruct,
perhaps due to the problems being too difficult for models of this size.

\paragraph{\OurTechnique{} works with easier problems}

All of the models we discussed so far were trained on \OurDsCodeforces.
To show that \OurTechnique{} works with other datasets, we also train models for Julia, Lua, and R using the MBPP training set.
(\Cref{sec:preparation} describes how we prepare MBPP.)
MBPP problems are trivial compared to the Open-R1 Codeforces problems (see \cref{fig:mbpp-example}),
and we cannot expect models trained on the MBPP problems to be as good as ones we presented before.
Still, training on MBPP improves Lua and Julia performance (\cref{tab:results-lcbx-main,tab:results-multiple}).
The table shows a small drop in R performance on \OurDsLcb, but a significant improvement on MultiPL-E.

\begin{wraptable}{r}{.46\textwidth}
\vspace{-1\intextsep}
\footnotesize
\centering
\begin{booktabs}{colspec={@{}l|[gray8]rrr|[gray8]rrr@{}}, colsep=2pt}
  \toprule
  Model & \SetCell[c=3]{c} MultiPL-E & & & \SetCell[c=3]{c} Ag-LCB-X \\
  \SetCell{r} X= & Lua & Julia & R & Lua & Julia & R \\
  \midrule
  SmolLM3 3B
        & 11 & 12 & 18
        & 1 & 2 & 2 \\
  Phi 4 mini ins
        & 40 & 39 & 34
        & 8 & 8 & 5 \\
  DSC 6.7B Ins
        & 40 & 54 & 37
        & 8 & 5 & 8 \\
  \SetRow{azure9}
  SmolLM3-3B-CF-X
        & 14 & 14 & 21
        & 8 & 8 & 6 \\
  \SetRow{azure9}
  Phi4-mini-ins-CF-X
        & 41 & 43 & 35
        & \textbf{12} & 8 & \textbf{12} \\
  \SetRow{azure9}
  DSC-6.7B-Ins-CF-X
        & \textbf{42} & \textbf{55} & \textbf{52}
        & 9 & \textbf{9} & 10 \\
  \bottomrule
\end{booktabs}
\caption{
  Non-Qwen models, pass@1.
}
\label{tab:results-other}
\vspace{-1\intextsep}
\end{wraptable}

\paragraph{\OurTechnique{} works on multiple model families}
To show that \OurTechnique{} works on non-Qwen models,
we train
SmolLM~3~\citep{smollm3},
Phi~4 Mini Instruct~\citep{microsoft2025phi4minitechnicalreportcompact},
and DeepSeek Coder~6.7B Instruct~\citep{guo2024deepseekcoderlargelanguagemodel}
on \OurDsCodeforces\ specialized to Lua, Julia, and R.
\OurTechnique{} improves these models' performance on all languages,
as measured by MultiPL-E and \OurDsLcb{} (\Cref{tab:results-other}).

Note that DeepSeek Coder 6.7B is a relatively old LLM, superseded by the much larger DeepSeekV2 and V3 models. Unlike Qwen 3, DeepSeek Coder is not trained with reinforcement learning, but is only an instruction-tuned model. Thus this result also shows that \OurTechnique{} can work on models that have had relatively limited post-training.

\begin{wraptable}{r}{.45\textwidth}
\vspace{-1\intextsep}
\footnotesize
\centering
\begin{booktabs}{colspec={@{}l|[gray8]r@{}}, colsep=3pt}
\toprule
Model & Ag-LCB-Fortran \\
\midrule
Sonnet 4 Thinking (teacher)
      & 12 \\
Qwen 3 4B (student)
      & 0 \\
\SetRow{azure9}
1 epoch
      & 3 \\
\SetRow{azure9}
2 epochs
      & 3 \\
\SetRow{azure9}
3 epochs
      & 2 \\
\bottomrule
\end{booktabs}
\caption{Distillation experiment results.}
\label{tab:distillation}
\vspace{-1\intextsep}
\end{wraptable}

\paragraph{\OurTechnique\ outperforms distillation}
So far we discussed training a model on its generations.
An alternative is to distill a larger model (assuming one exists).
As larger models do not perform very well on many low-resource programming languages,
one can expect distillation to be less effective.

We run the following experiment to verify this claim.
Using Sonnet 4 Thinking,
we synthesize Fortran solutions to \OurDsCodeforces\ problems,
creating a training set of 1,987 items.
(For 13 items, Sonnet 4 (sonnet-4-20250514) with extended thinking does not produce a response within its reasoning budget.)
To make sure generating the training items does not use significantly more compute compared to \OurTechnique\ training,
we use at most 32K reasoning tokens, spending $\$96$ to generate the items.
We fine-tune Qwen 3 4B for 3 epochs (batch size 64, learning rate $2 \times 10^{-5}$, cosine learning rate decay with warmup ratio $0.1$).
\Cref{tab:distillation} shows the resulting models reach scores
far lower than the $15\%$ of Qwen3-4B-CF-Fortran (\cref{tab:results-lcbx-aux}).

\paragraph{Agnostics outperforms rejection sampling}
\label{sec:rejection-sampling-efficiency}
We also ran a small experiment to confirm that rejection sampling would be prohibitively expensive in our case.
In rejection sampling with supervised fine-tuning ,
we prompt the model to synthesize $n$ solutions to each task, reject solutions that fail tests,
and fine-tune on the task-solution pairs that pass tests.
\Citet{cassano2024} use this approach to get solutions for ${\approx} 30\%$ of the tasks in their dataset.

The efficiency of this approach depends on the hardness of the task and the capabilities of the model.
In this paper, we use newer models that are marginally better at low-resource languages (based on MultiPL-E benchmark results).
The task of \citet{cassano2024} is to translate a simple, self-contained Python function from the model's pretraining data into an equivalent function in another programming language.
This task is significantly easier than the \OurDsCodeforces\ task,
which is to solve a competitive programming problem in a low-resource language without any reference code.

Rejection sampling would be prohibitively expensive for the low-resource programming languages we consider.
\Citet{cassano2024} report a $30\%$ success rate on their code translation task.
During \OurTechnique\ training, Qwen3-4B-CF-Fortran generated a correct answer to a train split problem only $6.64\%$ of the time,
generating $11400$ verified programs overall.
We also sampled responses to the same problems from the base model of Qwen3-4B-CF-Fortran, Qwen 3 4B,
taking the same amount of samples with the same generation parameters as used during training.
The base model succeeded $0.09\%$ of the time, generating only $158$ test-passing programs.

\begin{wrapfigure}{R}{.5\textwidth}
\vspace{-1\intextsep}
\centering
\includegraphics[width=0.5\textwidth]{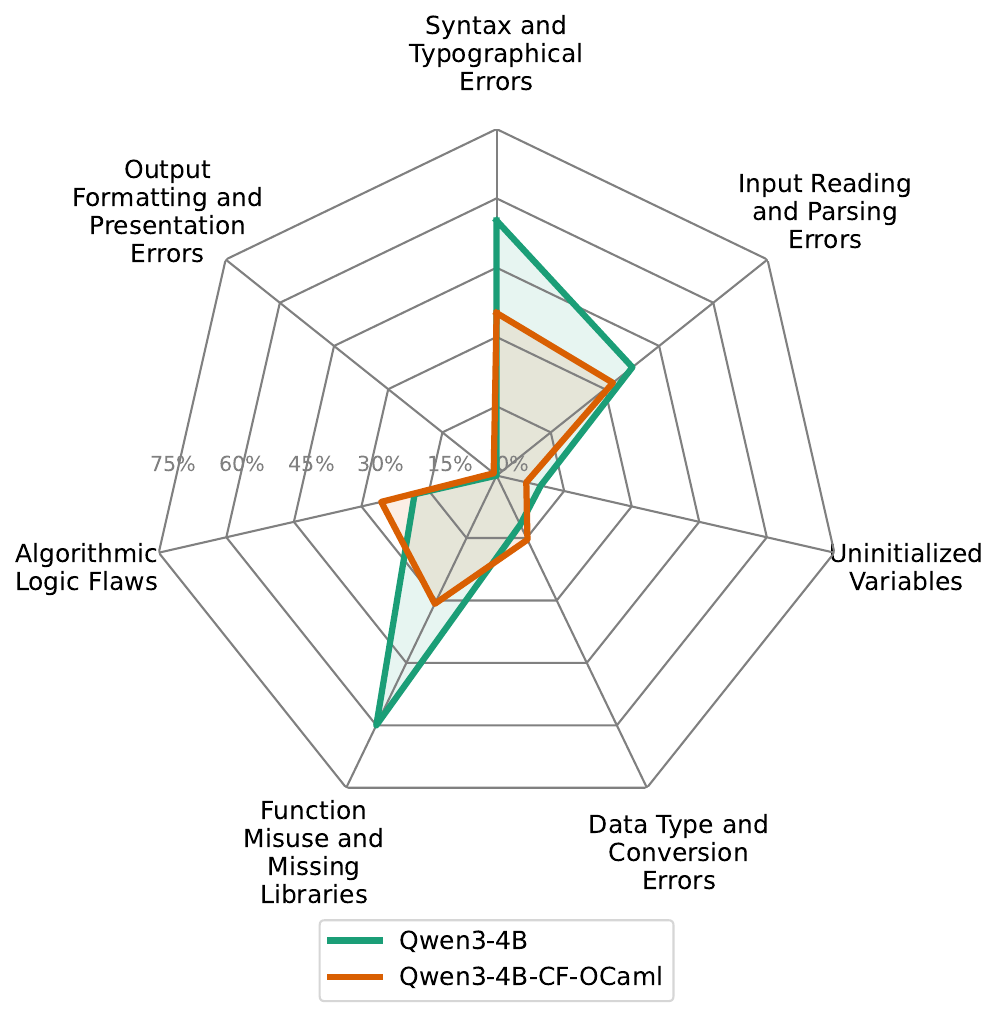}
\caption{
  LLM labels of bugs in programs
  synthesized by Qwen 3 4B and our trained model, Qwen3-4B-CF-OCaml.
  A radial represents a bug class,
  and points along the radial show how many programs seem to have that bug.
  Qwen 3 4B makes more fundamental programming mistakes,
  such as syntax errors and misusing builtins,
  showing its limited grasp of OCaml.
  Our trained model makes far fewer such mistakes.
  We see a small increase in logic flaws:
  the trained model makes fewer shallow mistakes, revealing deeper issues.
}
\label{fig:radar-ml}
\vspace{-2\intextsep}
\end{wrapfigure}

\subsection{Qualitative Improvements}

We now take a deeper look at how training with \OurTechnique\
qualitatively improves Qwen 3 4B.
First, we define a taxonomy of common bugs;
we prompt o3 to classify bugs in a sample of faulty programs
and lightly edit the suggestion.
\Cref{appendix:app-error-taxonomy} has the full taxonomy and the prompt used to develop it.
The taxonomy spans fundamental programming errors such as syntax errors,
and subtler mistakes such as logic flaws.

We sample 100 problems from \OurDsLcb{},
and for each take five OCaml programs produced by Qwen 3 4B and
the models trained on \OurDsCodeforces.
Recall from \cref{tab:results-lcbx-aux} that our models significantly outperform the base Qwen 3 models on our benchmark.
With Sonnet 4, we use the taxonomy to classify the bugs in each program.

\Cref{fig:radar-ml} shows the bug distribution for OCaml.
We see that the base Qwen 3 4B model makes many more fundamental OCaml programming mistakes.
More programs have syntax errors (55\% vs 35\% after training),
more programs misuse builtin functions (60\% vs 32\%), and so on.
We do observe a small increase in logic flaws (18\% vs 25\% after training).
When a program is full of syntax errors and hallucinated functions,
it is hard to decide if the algorithmic approach is correct;
training eliminates these shallow bugs and lets deeper issues manifest.
Models trained for the other four programming languages show the same patterns (\cref{appendix:app-error-taxonomy}).

\section{Conclusion}

Arguably, practicioners need LLMs the most for low-resource programming languages,
which need highly specialized knowledge.
Unfortunately, this is precisely where LLMs are weak due to a lack of sufficient pre-training and post-training data.
We propose \OurTechnique{},
a language-agnostic post-training pipeline that minimizes the per-language engineering tax by verifying code entirely via externally observable behavior.
A short configuration file is enough to adapt the pipeline, including the reinforcement learning setup, to a new programming language.

Empirically, \OurTechnique{} consistently improves small open-weight models on five low-resource languages%
---Lua, Julia, R, OCaml, and Fortran---without requiring language-specific test translators.
Training Qwen 3 4B with \OurDsCodeforces{} yields large gains on our new \OurDsLcb{} benchmark and on MultiPL-E,
often rivaling or surpassing 16B--70B open-weight baselines.
The method scales to larger and different model families:
Qwen 3 8B shows similar gains to its smaller sibling,
and we also observe improvements on DeepSeek Coder 6.7B, Phi 4 Mini and SmolLM3 3B.
Error-type analysis shows our training decreases fundamental programming language mistakes.

A practical advantage of the approach is how little per-language work it requires.
After the framework was in place, extending the pipeline to support OCaml and Fortran took us less than an hour each.
We expect adaption to be just as straightforward for any pragmatic programming language with a command-line toolchain.

We believe the approach scales to models of arbitrary size,
although our experiments are limited by available compute to at most 8B models.
For scaling data, the \OurTechnique\ reformulation approach also applies to much larger problem sets.
For instance, OpenCodeReasoning has $\sim$600K problems with Python solutions \citep{ahmad2025opencodereasoning};
converting such corpora into language-agnostic I/O tasks would provide rich RL datasets with many target languages with minimal additional engineering.

\section*{Reproducibility Statement}
\label{sec:reproducibility}

We provide references to our data, code, and artifacts at \href{https://agnostics.abgru.me}{agnostics.abgru.me}.
We share
our datasets and models at \href{https://huggingface.co/collections/nuprl/agnostics}{huggingface.co/collections/nuprl/agnostics},
our Wandb training logs at \href{https://wandb.ai/nuprl/Agnostics}{wandb.ai/nuprl/Agnostics},
and our training framework at \href{https://github.com/nuprl/agnostics-framework}{github.com/nuprl/agnostics-framework}.

The existing datasets we used are publicly available and are accompanied by citations.
We publicly release all datasets we introduce, allowing free use for research.
In the same way, we release
all code required for conducting and analyzing our experiments,
including the code for dataset preparation,
as well as the models we presented and Wandb records from training them.
We state the number and range of values tried per (hyper) parameter,
and outline how we chose the final values
and what they are (\cref{appendix:choosing-hyperparameters,sec:hyperparameters}).
We specify the computing infrastructure (hardware and software) we used for our experiments (\cref{appendix:used-hardware-software}).
The released codebases specify the exact versions of all the libraries we used.

\ifanon
\else
\section*{Acknowledgments}
This work is partially supported by the
U.S. National Science Foundation (SES-2326174). This material is based upon
work supported by the U.S. Department of Energy, Office of Science under Award Number DESC0025613.

This research used resources of the National Energy Research Scientific Computing Center (NERSC),
a Department of Energy User Facility using NERSC award ALCC-ERCAP 0038272 (m5083-2024).

We thank Northeastern Research Computing for support with the Northeastern University Explorer cluster. This work used the Delta cluster at the National Center for Supercomputing Applications (NCSA) through allocation CIS230213 from the Advanced Cyberinfrastructure Coordination Ecosystem: Services \& Support (ACCESS) program, which is supported by U.S. National Science Foundation grants 2138259, 2138286, 2138307, 2137603, and 2138296.

This work was supported by MEYS, ERC CZ program, grant no. LL2325.

This research was funded in part by the Swiss National Science Foundation (SNSF), Postdoc.Mobility grant P500PT\_225421.

This work uses vLLM~\citep{kwon2023efficient}, Transformers~\citep{wolf-etal-2020-transformers}, and GNU Parallel~\citep{tange2023}.

\emph{Disclaimer}: This report was prepared as an
account of work sponsored by an agency of the
United States Government. Neither the United
States Government nor any agency thereof, nor
any of their employees, makes any warranty, express or implied, or assumes any legal liability
or responsibility for the accuracy, completeness,
or usefulness of any information, apparatus, product, or process disclosed, or represents that its use
would not infringe privately owned rights. Reference herein to any specific commercial product,
process, or service by trade name, trademark, manufacturer, or otherwise does not necessarily constitute or imply its endorsement, recommendation,
or favoring by the United States Government or
any agency thereof. The views and opinions of
authors expressed herein do not necessarily state
or reflect those of the United States Government or any agency thereof.
\fi

\bibliography{autogenerated-abg,main}

\begin{thebibliography}{47}
\providecommand{\natexlab}[1]{#1}
\providecommand{\url}[1]{\texttt{#1}}
\expandafter\ifx\csname urlstyle\endcsname\relax
  \providecommand{\doi}[1]{doi: #1}\else
  \providecommand{\doi}{doi: \begingroup \urlstyle{rm}\Url}\fi

\bibitem[Ahmad et~al.(2025)Ahmad, Sean~Narenthiran, Ficek, Jain, Huang,
  Noroozi, and Ginsburg]{ahmad2025opencodereasoning}
Wasi~Uddin Ahmad, Somshubra~Majumdar Sean~Narenthiran, Aleksander Ficek,
  Siddhartha Jain, Jocelyn Huang, Vahid Noroozi, and Boris Ginsburg.
\newblock Opencodereasoning: Advancing data distillation for competitive
  coding, 2025.
\newblock URL \url{https://arxiv.org/abs/2504.01943}.

\bibitem[Allal et~al.(2023)Allal, Li, Kocetkov, Mou, Akiki, Ferrandis,
  Muennighoff, Mishra, Gu, Dey, Umapathi, Anderson, Zi, Poirier, Schoelkopf,
  Troshin, Abulkhanov, Romero, Lappert, De~Toni, {del R{\'i}o}, Liu, Bose,
  Bhattacharyya, Zhuo, Yu, Villegas, Zocca, Mangrulkar, Lansky, Nguyen,
  Contractor, Villa, Li, Bahdanau, Jernite, Hughes, Fried, Guha, {de Vries},
  and {von Werra}]{allal:santacoder}
Loubna~Ben Allal, Raymond Li, Denis Kocetkov, Chenghao Mou, Christopher Akiki,
  Carlos~Munoz Ferrandis, Niklas Muennighoff, Mayank Mishra, Alex Gu, Manan
  Dey, Logesh~Kumar Umapathi, Carolyn~Jane Anderson, Yangtian Zi, Joel~Lamy
  Poirier, Hailey Schoelkopf, Sergey Troshin, Dmitry Abulkhanov, Manuel Romero,
  Michael Lappert, Francesco De~Toni, Bernardo~Garc{\'i}a {del R{\'i}o}, Qian
  Liu, Shamik Bose, Urvashi Bhattacharyya, Terry~Yue Zhuo, Ian Yu, Paulo
  Villegas, Marco Zocca, Sourab Mangrulkar, David Lansky, Huu Nguyen, Danish
  Contractor, Luis Villa, Jia Li, Dzmitry Bahdanau, Yacine Jernite, Sean
  Hughes, Daniel Fried, Arjun Guha, Harm {de Vries}, and Leandro {von Werra}.
\newblock {{SantaCoder}}: Don't reach for the stars!
\newblock In \emph{Deep {{Learning}} for {{Code Workshop}} ({{DL4C}})}, 2023.

\bibitem[Austin et~al.(2021)Austin, Odena, Nye, Bosma, Michalewski, Dohan,
  Jiang, Cai, Terry, Le, and Sutton]{austin2021}
Jacob Austin, Augustus Odena, Maxwell Nye, Maarten Bosma, Henryk Michalewski,
  David Dohan, Ellen Jiang, Carrie Cai, Michael Terry, Quoc Le, and Charles
  Sutton.
\newblock Program {{Synthesis}} with {{Large Language Models}}.
\newblock Technical Report arXiv:2108.07732, arXiv, August 2021.
\newblock URL \url{http://arxiv.org/abs/2108.07732}.

\bibitem[ByteDance et~al.(2025)ByteDance, Zhang, Su, Sun, Xi, Xiao, Zheng,
  Zhang, Liu, Zan, Sun, Zhu, Xin, Huang, Bai, Dong, Li, Chen, Zhou, Huang,
  Ning, Song, Chen, Liu, Shen, Xiang, and Wu]{yuyu:seedcoder}
ByteDance, Yuyu Zhang, Jing Su, Yifan Sun, Chenguang Xi, Xia Xiao, Shen Zheng,
  Anxiang Zhang, Kaibo Liu, Daoguang Zan, Tao Sun, Jinhua Zhu, Shulin Xin, Dong
  Huang, Yetao Bai, Lixin Dong, Chao Li, Jianchong Chen, Hanzhi Zhou, Yifan
  Huang, Guanghan Ning, Xierui Song, Jiaze Chen, Siyao Liu, Kai Shen, Liang
  Xiang, and Yonghui Wu.
\newblock Seed-{{Coder}}: {{Let}} the {{Code Model Curate Data}} for
  {{Itself}}, June 2025.

\bibitem[Cassano et~al.(2023)Cassano, Gouwar, Nguyen, Nguyen, {Phipps-Costin},
  Pinckney, Yee, Zi, Anderson, Feldman, Guha, Greenberg, and
  Jangda]{cassano:multipl-e}
Federico Cassano, John Gouwar, Daniel Nguyen, Sydney Nguyen, Luna
  {Phipps-Costin}, Donald Pinckney, Ming-Ho Yee, Yangtian Zi, Carolyn~Jane
  Anderson, Molly~Q. Feldman, Arjun Guha, Michael Greenberg, and Abhinav
  Jangda.
\newblock {{MultiPL-E}}: {{A Scalable}} and {{Polyglot Approach}} to
  {{Benchmarking Neural Code Generation}}.
\newblock \emph{IEEE Transactions on Software Engineering (TSE)}, 49\penalty0
  (7):\penalty0 3675--3691, 2023.

\bibitem[Cassano et~al.(2024)Cassano, Gouwar, Lucchetti, Schlesinger, Freeman,
  Anderson, Feldman, Greenberg, Jangda, and Guha]{cassano2024}
Federico Cassano, John Gouwar, Francesca Lucchetti, Claire Schlesinger, Anders
  Freeman, Carolyn~Jane Anderson, Molly~Q Feldman, Michael Greenberg, Abhinav
  Jangda, and Arjun Guha.
\newblock Knowledge {{Transfer}} from {{High-Resource}} to {{Low-Resource
  Programming Languages}} for {{Code LLMs}}.
\newblock \emph{Artifact: Knowledge Transfer from High-Resource to Low-Resource
  Programming Languages for Code LLMs}, 8\penalty0 (OOPSLA2):\penalty0
  295:677--295:708, October 2024.
\newblock \doi{10.1145/3689735}.
\newblock URL \url{https://dl.acm.org/doi/10.1145/3689735}.

\bibitem[Chen et~al.(2021)Chen, Tworek, Jun, Yuan, Pinto, Kaplan, Edwards,
  Burda, Joseph, Brockman, Ray, Puri, Krueger, Petrov, Khlaaf, Sastry, Mishkin,
  Chan, Gray, Ryder, Pavlov, Power, Kaiser, Bavarian, Winter, Tillet, Such,
  Cummings, Plappert, Chantzis, Barnes, {Herbert-Voss}, Guss, Nichol, Paino,
  Tezak, Tang, Babuschkin, Balaji, Jain, Saunders, Hesse, Carr, Leike, Achiam,
  Misra, Morikawa, Radford, Knight, Brundage, Murati, Mayer, Welinder, McGrew,
  Amodei, McCandlish, Sutskever, and Zaremba]{chen:codex}
Mark Chen, Jerry Tworek, Heewoo Jun, Qiming Yuan, Henrique Ponde de~Oliveira
  Pinto, Jared Kaplan, Harri Edwards, Yuri Burda, Nicholas Joseph, Greg
  Brockman, Alex Ray, Raul Puri, Gretchen Krueger, Michael Petrov, Heidy
  Khlaaf, Girish Sastry, Pamela Mishkin, Brooke Chan, Scott Gray, Nick Ryder,
  Mikhail Pavlov, Alethea Power, Lukasz Kaiser, Mohammad Bavarian, Clemens
  Winter, Philippe Tillet, Felipe~Petroski Such, Dave Cummings, Matthias
  Plappert, Fotios Chantzis, Elizabeth Barnes, Ariel {Herbert-Voss},
  William~Hebgen Guss, Alex Nichol, Alex Paino, Nikolas Tezak, Jie Tang, Igor
  Babuschkin, Suchir Balaji, Shantanu Jain, William Saunders, Christopher
  Hesse, Andrew~N. Carr, Jan Leike, Josh Achiam, Vedant Misra, Evan Morikawa,
  Alec Radford, Matthew Knight, Miles Brundage, Mira Murati, Katie Mayer, Peter
  Welinder, Bob McGrew, Dario Amodei, Sam McCandlish, Ilya Sutskever, and
  Wojciech Zaremba.
\newblock Evaluating {{Large Language Models Trained}} on {{Code}}, July 2021.

\bibitem[{Decon}(2025)]{decon}
{Decon}.
\newblock Decon.
\newblock \url{https://github.com/allenai/decon}, 2025.

\bibitem[{DeepSeek-AI} et~al.(2025){DeepSeek-AI}, Guo, Yang, Zhang, Song,
  Zhang, Xu, Zhu, Ma, Wang, Bi, Zhang, Yu, Wu, Wu, Gou, Shao, Li, Gao, Liu,
  Xue, Wang, Wu, Feng, Lu, Zhao, Deng, Zhang, Ruan, Dai, Chen, Ji, Li, Lin,
  Dai, Luo, Hao, Chen, Li, Zhang, Bao, Xu, Wang, Ding, Xin, Gao, Qu, Li, Guo,
  Li, Wang, Chen, Yuan, Qiu, Li, Cai, Ni, Liang, Chen, Dong, Hu, Gao, Guan,
  Huang, Yu, Wang, Zhang, Zhao, Wang, Zhang, Xu, Xia, Zhang, Zhang, Tang, Li,
  Wang, Li, Tian, Huang, Zhang, Wang, Chen, Du, Ge, Zhang, Pan, Wang, Chen,
  Jin, Chen, Lu, Zhou, Chen, Ye, Wang, Yu, Zhou, Pan, Li, Zhou, Wu, Ye, Yun,
  Pei, Sun, Wang, Zeng, Zhao, Liu, Liang, Gao, Yu, Zhang, Xiao, An, Liu, Wang,
  Chen, Nie, Cheng, Liu, Xie, Liu, Yang, Li, Su, Lin, Li, Jin, Shen, Chen, Sun,
  Wang, Song, Zhou, Wang, Shan, Li, Wang, Wei, Zhang, Xu, Li, Zhao, Sun, Wang,
  Yu, Zhang, Shi, Xiong, He, Piao, Wang, Tan, Ma, Liu, Guo, Ou, Wang, Gong,
  Zou, He, Xiong, Luo, You, Liu, Zhou, Zhu, Xu, Huang, Li, Zheng, Zhu, Ma,
  Tang, Zha, Yan, Ren, Ren, Sha, Fu, Xu, Xie, Zhang, Hao, Ma, Yan, Wu, Gu, Zhu,
  Liu, Li, Xie, Song, Pan, Huang, Xu, Zhang, and Zhang]{deepseek-ai2025}
{DeepSeek-AI}, Daya Guo, Dejian Yang, Haowei Zhang, Junxiao Song, Ruoyu Zhang,
  Runxin Xu, Qihao Zhu, Shirong Ma, Peiyi Wang, Xiao Bi, Xiaokang Zhang,
  Xingkai Yu, Yu~Wu, Z.~F. Wu, Zhibin Gou, Zhihong Shao, Zhuoshu Li, Ziyi Gao,
  Aixin Liu, Bing Xue, Bingxuan Wang, Bochao Wu, Bei Feng, Chengda Lu,
  Chenggang Zhao, Chengqi Deng, Chenyu Zhang, Chong Ruan, Damai Dai, Deli Chen,
  Dongjie Ji, Erhang Li, Fangyun Lin, Fucong Dai, Fuli Luo, Guangbo Hao,
  Guanting Chen, Guowei Li, H.~Zhang, Han Bao, Hanwei Xu, Haocheng Wang,
  Honghui Ding, Huajian Xin, Huazuo Gao, Hui Qu, Hui Li, Jianzhong Guo, Jiashi
  Li, Jiawei Wang, Jingchang Chen, Jingyang Yuan, Junjie Qiu, Junlong Li, J.~L.
  Cai, Jiaqi Ni, Jian Liang, Jin Chen, Kai Dong, Kai Hu, Kaige Gao, Kang Guan,
  Kexin Huang, Kuai Yu, Lean Wang, Lecong Zhang, Liang Zhao, Litong Wang, Liyue
  Zhang, Lei Xu, Leyi Xia, Mingchuan Zhang, Minghua Zhang, Minghui Tang, Meng
  Li, Miaojun Wang, Mingming Li, Ning Tian, Panpan Huang, Peng Zhang, Qiancheng
  Wang, Qinyu Chen, Qiushi Du, Ruiqi Ge, Ruisong Zhang, Ruizhe Pan, Runji Wang,
  R.~J. Chen, R.~L. Jin, Ruyi Chen, Shanghao Lu, Shangyan Zhou, Shanhuang Chen,
  Shengfeng Ye, Shiyu Wang, Shuiping Yu, Shunfeng Zhou, Shuting Pan, S.~S. Li,
  Shuang Zhou, Shaoqing Wu, Shengfeng Ye, Tao Yun, Tian Pei, Tianyu Sun,
  T.~Wang, Wangding Zeng, Wanjia Zhao, Wen Liu, Wenfeng Liang, Wenjun Gao,
  Wenqin Yu, Wentao Zhang, W.~L. Xiao, Wei An, Xiaodong Liu, Xiaohan Wang,
  Xiaokang Chen, Xiaotao Nie, Xin Cheng, Xin Liu, Xin Xie, Xingchao Liu, Xinyu
  Yang, Xinyuan Li, Xuecheng Su, Xuheng Lin, X.~Q. Li, Xiangyue Jin, Xiaojin
  Shen, Xiaosha Chen, Xiaowen Sun, Xiaoxiang Wang, Xinnan Song, Xinyi Zhou,
  Xianzu Wang, Xinxia Shan, Y.~K. Li, Y.~Q. Wang, Y.~X. Wei, Yang Zhang,
  Yanhong Xu, Yao Li, Yao Zhao, Yaofeng Sun, Yaohui Wang, Yi~Yu, Yichao Zhang,
  Yifan Shi, Yiliang Xiong, Ying He, Yishi Piao, Yisong Wang, Yixuan Tan,
  Yiyang Ma, Yiyuan Liu, Yongqiang Guo, Yuan Ou, Yuduan Wang, Yue Gong, Yuheng
  Zou, Yujia He, Yunfan Xiong, Yuxiang Luo, Yuxiang You, Yuxuan Liu, Yuyang
  Zhou, Y.~X. Zhu, Yanhong Xu, Yanping Huang, Yaohui Li, Yi~Zheng, Yuchen Zhu,
  Yunxian Ma, Ying Tang, Yukun Zha, Yuting Yan, Z.~Z. Ren, Zehui Ren, Zhangli
  Sha, Zhe Fu, Zhean Xu, Zhenda Xie, Zhengyan Zhang, Zhewen Hao, Zhicheng Ma,
  Zhigang Yan, Zhiyu Wu, Zihui Gu, Zijia Zhu, Zijun Liu, Zilin Li, Ziwei Xie,
  Ziyang Song, Zizheng Pan, Zhen Huang, Zhipeng Xu, Zhongyu Zhang, and Zhen
  Zhang.
\newblock {{DeepSeek-R1}}: {{Incentivizing Reasoning Capability}} in {{LLMs}}
  via {{Reinforcement Learning}}.
\newblock Technical Report arXiv:2501.12948, arXiv, January 2025.
\newblock URL \url{http://arxiv.org/abs/2501.12948}.

\bibitem[Du et~al.(2025)Du, Tuan, Liu, Qing, Huang, He, Liu, Ma, and
  Ng]{du:afterburner}
Mingzhe Du, Luu~Anh Tuan, Yue Liu, Yuhao Qing, Dong Huang, Xinyi He, Qian Liu,
  Zejun Ma, and See-kiong Ng.
\newblock Afterburner: {{Reinforcement Learning Facilitates Self-Improving Code
  Efficiency Optimization}}, June 2025.

\bibitem[{Firefox}(2025)]{firefox-ramdisk}
{Firefox}.
\newblock Why the build system might be slow.
\newblock
  \url{https://firefox-source-docs.mozilla.org/build/buildsystem/slow.html},
  2025.

\bibitem[Gehring et~al.(2024)Gehring, Zheng, Copet, Mella, Cohen, and
  Synnaeve]{gehring:rlef}
Jonas Gehring, Kunhao Zheng, Jade Copet, Vegard Mella, Taco Cohen, and Gabriel
  Synnaeve.
\newblock {{RLEF}}: {{Grounding Code LLMs}} in {{Execution Feedback}} with
  {{Reinforcement Learning}}, October 2024.

\bibitem[Grattafiori et~al.(2024)Grattafiori, Dubey, Jauhri, Pandey, Kadian,
  {Al-Dahle}, Letman, Mathur, Schelten, Vaughan, Yang, Fan, Goyal, Hartshorn,
  Yang, Mitra, Sravankumar, Korenev, Hinsvark, Rao, Zhang, Rodriguez,
  Gregerson, Spataru, Roziere, Biron, Tang, Chern, Caucheteux, Nayak, Bi,
  Marra, McConnell, Keller, Touret, Wu, Wong, Ferrer, Nikolaidis, Allonsius,
  Song, Pintz, Livshits, Wyatt, Esiobu, Choudhary, Mahajan, {Garcia-Olano},
  Perino, Hupkes, Lakomkin, AlBadawy, Lobanova, Dinan, Smith, Radenovic,
  Guzm{\'a}n, Zhang, Synnaeve, Lee, Anderson, Thattai, Nail, Mialon, Pang,
  Cucurell, Nguyen, Korevaar, Xu, Touvron, Zarov, Ibarra, Kloumann, Misra,
  Evtimov, Zhang, Copet, Lee, Geffert, Vranes, Park, Mahadeokar, Shah, van~der
  Linde, Billock, Hong, Lee, Fu, Chi, Huang, Liu, Wang, Yu, Bitton, Spisak,
  Park, Rocca, Johnstun, Saxe, Jia, Alwala, Prasad, Upasani, Plawiak, Li,
  Heafield, Stone, {El-Arini}, Iyer, Malik, Chiu, Bhalla, Lakhotia,
  {Rantala-Yeary}, van~der Maaten, Chen, Tan, Jenkins, Martin, Madaan, Malo,
  Blecher, Landzaat, de~Oliveira, Muzzi, Pasupuleti, Singh, Paluri, Kardas,
  Tsimpoukelli, Oldham, Rita, Pavlova, Kambadur, Lewis, Si, Singh, Hassan,
  Goyal, Torabi, Bashlykov, Bogoychev, Chatterji, Zhang, Duchenne, {\c C}elebi,
  Alrassy, Zhang, Li, Vasic, Weng, Bhargava, Dubal, Krishnan, Koura, Xu, He,
  Dong, Srinivasan, Ganapathy, Calderer, Cabral, Stojnic, Raileanu, Maheswari,
  Girdhar, Patel, Sauvestre, Polidoro, Sumbaly, Taylor, Silva, Hou, Wang,
  Hosseini, Chennabasappa, Singh, Bell, Kim, Edunov, Nie, Narang, Raparthy,
  Shen, Wan, Bhosale, Zhang, Vandenhende, Batra, Whitman, Sootla, Collot,
  Gururangan, Borodinsky, Herman, Fowler, Sheasha, Georgiou, Scialom,
  Speckbacher, Mihaylov, Xiao, Karn, Goswami, Gupta, Ramanathan, Kerkez,
  Gonguet, Do, Vogeti, Albiero, Petrovic, Chu, Xiong, Fu, Meers, Martinet,
  Wang, Wang, Tan, Xia, Xie, Jia, Wang, Goldschlag, Gaur, Babaei, Wen, Song,
  Zhang, Li, Mao, Coudert, Yan, Chen, Papakipos, Singh, Srivastava, Jain,
  Kelsey, Shajnfeld, Gangidi, Victoria, Goldstand, Menon, Sharma, Boesenberg,
  Baevski, Feinstein, Kallet, Sangani, Teo, Yunus, Lupu, Alvarado, Caples, Gu,
  Ho, Poulton, Ryan, Ramchandani, Dong, Franco, Goyal, Saraf, Chowdhury,
  Gabriel, Bharambe, Eisenman, Yazdan, James, Maurer, Leonhardi, Huang, Loyd,
  Paola, Paranjape, Liu, Wu, Ni, Hancock, Wasti, Spence, Stojkovic, Gamido,
  Montalvo, Parker, Burton, Mejia, Liu, Wang, Kim, Zhou, Hu, Chu, Cai, Tindal,
  Feichtenhofer, Gao, Civin, Beaty, Kreymer, Li, Adkins, Xu, Testuggine, David,
  Parikh, Liskovich, Foss, Wang, Le, Holland, Dowling, Jamil, Montgomery,
  Presani, Hahn, Wood, Le, Brinkman, Arcaute, Dunbar, Smothers, Sun, Kreuk,
  Tian, Kokkinos, Ozgenel, Caggioni, Kanayet, Seide, Florez, Schwarz, Badeer,
  Swee, Halpern, Herman, Sizov, Guangyi, Zhang, Lakshminarayanan, Inan,
  Shojanazeri, Zou, Wang, Zha, Habeeb, Rudolph, Suk, Aspegren, Goldman, Zhan,
  Damlaj, Molybog, Tufanov, Leontiadis, Veliche, Gat, Weissman, Geboski, Kohli,
  Lam, Asher, Gaya, Marcus, Tang, Chan, Zhen, Reizenstein, Teboul, Zhong, Jin,
  Yang, Cummings, Carvill, Shepard, McPhie, Torres, Ginsburg, Wang, Wu, U,
  Saxena, Khandelwal, Zand, Matosich, Veeraraghavan, Michelena, Li, Jagadeesh,
  Huang, Chawla, Huang, Chen, Garg, A, Silva, Bell, Zhang, Guo, Yu, Moshkovich,
  Wehrstedt, Khabsa, Avalani, Bhatt, Mankus, Hasson, Lennie, Reso, Groshev,
  Naumov, Lathi, Keneally, Liu, Seltzer, Valko, Restrepo, Patel, Vyatskov,
  Samvelyan, Clark, Macey, Wang, Hermoso, Metanat, Rastegari, Bansal,
  Santhanam, Parks, White, Bawa, Singhal, Egebo, Usunier, Mehta, Laptev, Dong,
  Cheng, Chernoguz, Hart, Salpekar, Kalinli, Kent, Parekh, Saab, Balaji,
  Rittner, Bontrager, Roux, Dollar, Zvyagina, Ratanchandani, Yuvraj, Liang,
  Alao, Rodriguez, Ayub, Murthy, Nayani, Mitra, Parthasarathy, Li, Hogan,
  Battey, Wang, Howes, Rinott, Mehta, Siby, Bondu, Datta, Chugh, Hunt, Dhillon,
  Sidorov, Pan, Mahajan, Verma, Yamamoto, Ramaswamy, Lindsay, Lindsay, Feng,
  Lin, Zha, Patil, Shankar, Zhang, Zhang, Wang, Agarwal, Sajuyigbe, Chintala,
  Max, Chen, Kehoe, Satterfield, Govindaprasad, Gupta, Deng, Cho, Virk,
  Subramanian, Choudhury, Goldman, Remez, Glaser, Best, Koehler, Robinson, Li,
  Zhang, Matthews, Chou, Shaked, Vontimitta, Ajayi, Montanez, Mohan, Kumar,
  Mangla, Ionescu, Poenaru, Mihailescu, Ivanov, Li, Wang, Jiang, Bouaziz,
  Constable, Tang, Wu, Wang, Wu, Gao, Kleinman, Chen, Hu, Jia, Qi, Li, Zhang,
  Zhang, Adi, Nam, Yu, Wang, Zhao, Hao, Qian, Li, He, Rait, DeVito, Rosnbrick,
  Wen, Yang, Zhao, and Ma]{llama3}
Aaron Grattafiori, Abhimanyu Dubey, Abhinav Jauhri, Abhinav Pandey, Abhishek
  Kadian, Ahmad {Al-Dahle}, Aiesha Letman, Akhil Mathur, Alan Schelten, Alex
  Vaughan, Amy Yang, Angela Fan, Anirudh Goyal, Anthony Hartshorn, Aobo Yang,
  Archi Mitra, Archie Sravankumar, Artem Korenev, Arthur Hinsvark, Arun Rao,
  Aston Zhang, Aurelien Rodriguez, Austen Gregerson, Ava Spataru, Baptiste
  Roziere, Bethany Biron, Binh Tang, Bobbie Chern, Charlotte Caucheteux, Chaya
  Nayak, Chloe Bi, Chris Marra, Chris McConnell, Christian Keller, Christophe
  Touret, Chunyang Wu, Corinne Wong, Cristian~Canton Ferrer, Cyrus Nikolaidis,
  Damien Allonsius, Daniel Song, Danielle Pintz, Danny Livshits, Danny Wyatt,
  David Esiobu, Dhruv Choudhary, Dhruv Mahajan, Diego {Garcia-Olano}, Diego
  Perino, Dieuwke Hupkes, Egor Lakomkin, Ehab AlBadawy, Elina Lobanova, Emily
  Dinan, Eric~Michael Smith, Filip Radenovic, Francisco Guzm{\'a}n, Frank
  Zhang, Gabriel Synnaeve, Gabrielle Lee, Georgia~Lewis Anderson, Govind
  Thattai, Graeme Nail, Gregoire Mialon, Guan Pang, Guillem Cucurell, Hailey
  Nguyen, Hannah Korevaar, Hu~Xu, Hugo Touvron, Iliyan Zarov, Imanol~Arrieta
  Ibarra, Isabel Kloumann, Ishan Misra, Ivan Evtimov, Jack Zhang, Jade Copet,
  Jaewon Lee, Jan Geffert, Jana Vranes, Jason Park, Jay Mahadeokar, Jeet Shah,
  Jelmer van~der Linde, Jennifer Billock, Jenny Hong, Jenya Lee, Jeremy Fu,
  Jianfeng Chi, Jianyu Huang, Jiawen Liu, Jie Wang, Jiecao Yu, Joanna Bitton,
  Joe Spisak, Jongsoo Park, Joseph Rocca, Joshua Johnstun, Joshua Saxe, Junteng
  Jia, Kalyan~Vasuden Alwala, Karthik Prasad, Kartikeya Upasani, Kate Plawiak,
  Ke~Li, Kenneth Heafield, Kevin Stone, Khalid {El-Arini}, Krithika Iyer,
  Kshitiz Malik, Kuenley Chiu, Kunal Bhalla, Kushal Lakhotia, Lauren
  {Rantala-Yeary}, Laurens van~der Maaten, Lawrence Chen, Liang Tan, Liz
  Jenkins, Louis Martin, Lovish Madaan, Lubo Malo, Lukas Blecher, Lukas
  Landzaat, Luke de~Oliveira, Madeline Muzzi, Mahesh Pasupuleti, Mannat Singh,
  Manohar Paluri, Marcin Kardas, Maria Tsimpoukelli, Mathew Oldham, Mathieu
  Rita, Maya Pavlova, Melanie Kambadur, Mike Lewis, Min Si, Mitesh~Kumar Singh,
  Mona Hassan, Naman Goyal, Narjes Torabi, Nikolay Bashlykov, Nikolay
  Bogoychev, Niladri Chatterji, Ning Zhang, Olivier Duchenne, Onur {\c C}elebi,
  Patrick Alrassy, Pengchuan Zhang, Pengwei Li, Petar Vasic, Peter Weng,
  Prajjwal Bhargava, Pratik Dubal, Praveen Krishnan, Punit~Singh Koura, Puxin
  Xu, Qing He, Qingxiao Dong, Ragavan Srinivasan, Raj Ganapathy, Ramon
  Calderer, Ricardo~Silveira Cabral, Robert Stojnic, Roberta Raileanu, Rohan
  Maheswari, Rohit Girdhar, Rohit Patel, Romain Sauvestre, Ronnie Polidoro,
  Roshan Sumbaly, Ross Taylor, Ruan Silva, Rui Hou, Rui Wang, Saghar Hosseini,
  Sahana Chennabasappa, Sanjay Singh, Sean Bell, Seohyun~Sonia Kim, Sergey
  Edunov, Shaoliang Nie, Sharan Narang, Sharath Raparthy, Sheng Shen, Shengye
  Wan, Shruti Bhosale, Shun Zhang, Simon Vandenhende, Soumya Batra, Spencer
  Whitman, Sten Sootla, Stephane Collot, Suchin Gururangan, Sydney Borodinsky,
  Tamar Herman, Tara Fowler, Tarek Sheasha, Thomas Georgiou, Thomas Scialom,
  Tobias Speckbacher, Todor Mihaylov, Tong Xiao, Ujjwal Karn, Vedanuj Goswami,
  Vibhor Gupta, Vignesh Ramanathan, Viktor Kerkez, Vincent Gonguet, Virginie
  Do, Vish Vogeti, V{\'i}tor Albiero, Vladan Petrovic, Weiwei Chu, Wenhan
  Xiong, Wenyin Fu, Whitney Meers, Xavier Martinet, Xiaodong Wang, Xiaofang
  Wang, Xiaoqing~Ellen Tan, Xide Xia, Xinfeng Xie, Xuchao Jia, Xuewei Wang,
  Yaelle Goldschlag, Yashesh Gaur, Yasmine Babaei, Yi~Wen, Yiwen Song, Yuchen
  Zhang, Yue Li, Yuning Mao, Zacharie~Delpierre Coudert, Zheng Yan, Zhengxing
  Chen, Zoe Papakipos, Aaditya Singh, Aayushi Srivastava, Abha Jain, Adam
  Kelsey, Adam Shajnfeld, Adithya Gangidi, Adolfo Victoria, Ahuva Goldstand,
  Ajay Menon, Ajay Sharma, Alex Boesenberg, Alexei Baevski, Allie Feinstein,
  Amanda Kallet, Amit Sangani, Amos Teo, Anam Yunus, Andrei Lupu, Andres
  Alvarado, Andrew Caples, Andrew Gu, Andrew Ho, Andrew Poulton, Andrew Ryan,
  Ankit Ramchandani, Annie Dong, Annie Franco, Anuj Goyal, Aparajita Saraf,
  Arkabandhu Chowdhury, Ashley Gabriel, Ashwin Bharambe, Assaf Eisenman, Azadeh
  Yazdan, Beau James, Ben Maurer, Benjamin Leonhardi, Bernie Huang, Beth Loyd,
  Beto~De Paola, Bhargavi Paranjape, Bing Liu, Bo~Wu, Boyu Ni, Braden Hancock,
  Bram Wasti, Brandon Spence, Brani Stojkovic, Brian Gamido, Britt Montalvo,
  Carl Parker, Carly Burton, Catalina Mejia, Ce~Liu, Changhan Wang, Changkyu
  Kim, Chao Zhou, Chester Hu, Ching-Hsiang Chu, Chris Cai, Chris Tindal,
  Christoph Feichtenhofer, Cynthia Gao, Damon Civin, Dana Beaty, Daniel
  Kreymer, Daniel Li, David Adkins, David Xu, Davide Testuggine, Delia David,
  Devi Parikh, Diana Liskovich, Didem Foss, Dingkang Wang, Duc Le, Dustin
  Holland, Edward Dowling, Eissa Jamil, Elaine Montgomery, Eleonora Presani,
  Emily Hahn, Emily Wood, Eric-Tuan Le, Erik Brinkman, Esteban Arcaute, Evan
  Dunbar, Evan Smothers, Fei Sun, Felix Kreuk, Feng Tian, Filippos Kokkinos,
  Firat Ozgenel, Francesco Caggioni, Frank Kanayet, Frank Seide,
  Gabriela~Medina Florez, Gabriella Schwarz, Gada Badeer, Georgia Swee, Gil
  Halpern, Grant Herman, Grigory Sizov, Guangyi, Zhang, Guna Lakshminarayanan,
  Hakan Inan, Hamid Shojanazeri, Han Zou, Hannah Wang, Hanwen Zha, Haroun
  Habeeb, Harrison Rudolph, Helen Suk, Henry Aspegren, Hunter Goldman, Hongyuan
  Zhan, Ibrahim Damlaj, Igor Molybog, Igor Tufanov, Ilias Leontiadis,
  Irina-Elena Veliche, Itai Gat, Jake Weissman, James Geboski, James Kohli,
  Janice Lam, Japhet Asher, Jean-Baptiste Gaya, Jeff Marcus, Jeff Tang,
  Jennifer Chan, Jenny Zhen, Jeremy Reizenstein, Jeremy Teboul, Jessica Zhong,
  Jian Jin, Jingyi Yang, Joe Cummings, Jon Carvill, Jon Shepard, Jonathan
  McPhie, Jonathan Torres, Josh Ginsburg, Junjie Wang, Kai Wu, Kam~Hou U, Karan
  Saxena, Kartikay Khandelwal, Katayoun Zand, Kathy Matosich, Kaushik
  Veeraraghavan, Kelly Michelena, Keqian Li, Kiran Jagadeesh, Kun Huang, Kunal
  Chawla, Kyle Huang, Lailin Chen, Lakshya Garg, Lavender A, Leandro Silva, Lee
  Bell, Lei Zhang, Liangpeng Guo, Licheng Yu, Liron Moshkovich, Luca Wehrstedt,
  Madian Khabsa, Manav Avalani, Manish Bhatt, Martynas Mankus, Matan Hasson,
  Matthew Lennie, Matthias Reso, Maxim Groshev, Maxim Naumov, Maya Lathi,
  Meghan Keneally, Miao Liu, Michael~L. Seltzer, Michal Valko, Michelle
  Restrepo, Mihir Patel, Mik Vyatskov, Mikayel Samvelyan, Mike Clark, Mike
  Macey, Mike Wang, Miquel~Jubert Hermoso, Mo~Metanat, Mohammad Rastegari,
  Munish Bansal, Nandhini Santhanam, Natascha Parks, Natasha White, Navyata
  Bawa, Nayan Singhal, Nick Egebo, Nicolas Usunier, Nikhil Mehta,
  Nikolay~Pavlovich Laptev, Ning Dong, Norman Cheng, Oleg Chernoguz, Olivia
  Hart, Omkar Salpekar, Ozlem Kalinli, Parkin Kent, Parth Parekh, Paul Saab,
  Pavan Balaji, Pedro Rittner, Philip Bontrager, Pierre Roux, Piotr Dollar,
  Polina Zvyagina, Prashant Ratanchandani, Pritish Yuvraj, Qian Liang, Rachad
  Alao, Rachel Rodriguez, Rafi Ayub, Raghotham Murthy, Raghu Nayani, Rahul
  Mitra, Rangaprabhu Parthasarathy, Raymond Li, Rebekkah Hogan, Robin Battey,
  Rocky Wang, Russ Howes, Ruty Rinott, Sachin Mehta, Sachin Siby, Sai~Jayesh
  Bondu, Samyak Datta, Sara Chugh, Sara Hunt, Sargun Dhillon, Sasha Sidorov,
  Satadru Pan, Saurabh Mahajan, Saurabh Verma, Seiji Yamamoto, Sharadh
  Ramaswamy, Shaun Lindsay, Shaun Lindsay, Sheng Feng, Shenghao Lin,
  Shengxin~Cindy Zha, Shishir Patil, Shiva Shankar, Shuqiang Zhang, Shuqiang
  Zhang, Sinong Wang, Sneha Agarwal, Soji Sajuyigbe, Soumith Chintala,
  Stephanie Max, Stephen Chen, Steve Kehoe, Steve Satterfield, Sudarshan
  Govindaprasad, Sumit Gupta, Summer Deng, Sungmin Cho, Sunny Virk, Suraj
  Subramanian, Sy~Choudhury, Sydney Goldman, Tal Remez, Tamar Glaser, Tamara
  Best, Thilo Koehler, Thomas Robinson, Tianhe Li, Tianjun Zhang, Tim Matthews,
  Timothy Chou, Tzook Shaked, Varun Vontimitta, Victoria Ajayi, Victoria
  Montanez, Vijai Mohan, Vinay~Satish Kumar, Vishal Mangla, Vlad Ionescu, Vlad
  Poenaru, Vlad~Tiberiu Mihailescu, Vladimir Ivanov, Wei Li, Wenchen Wang,
  Wenwen Jiang, Wes Bouaziz, Will Constable, Xiaocheng Tang, Xiaojian Wu,
  Xiaolan Wang, Xilun Wu, Xinbo Gao, Yaniv Kleinman, Yanjun Chen, Ye~Hu,
  Ye~Jia, Ye~Qi, Yenda Li, Yilin Zhang, Ying Zhang, Yossi Adi, Youngjin Nam,
  Yu, Wang, Yu~Zhao, Yuchen Hao, Yundi Qian, Yunlu Li, Yuzi He, Zach Rait,
  Zachary DeVito, Zef Rosnbrick, Zhaoduo Wen, Zhenyu Yang, Zhiwei Zhao, and
  Zhiyu Ma.
\newblock The {{Llama}} 3 {{Herd}} of {{Models}}, November 2024.

\bibitem[Guo et~al.(2024)Guo, Zhu, Yang, Xie, Dong, Zhang, Chen, Bi, Wu, Li,
  Luo, Xiong, and Liang]{guo2024deepseekcoderlargelanguagemodel}
Daya Guo, Qihao Zhu, Dejian Yang, Zhenda Xie, Kai Dong, Wentao Zhang, Guanting
  Chen, Xiao Bi, Y.~Wu, Y.~K. Li, Fuli Luo, Yingfei Xiong, and Wenfeng Liang.
\newblock Deepseek-coder: When the large language model meets programming --
  the rise of code intelligence, 2024.
\newblock URL \url{https://arxiv.org/abs/2401.14196}.

\bibitem[Hu et~al.(2025)Hu, Li, Guha, and Biswas]{hu2025pre}
Zichao Hu, Junyi~Jessy Li, Arjun Guha, and Joydeep Biswas.
\newblock Robo-{{Instruct}}: {{Simulator-Augmented Instruction Alignment For
  Finetuning Code LLMs}}, April 2025.
\newblock URL \url{http://arxiv.org/abs/2405.20179}.

\bibitem[Jain et~al.(2025)Jain, {Gonzalez-Pumariega}, Chen, Rush, Zhao, and
  Choudhury]{jain:mu-code}
Arnav~Kumar Jain, Gonzalo {Gonzalez-Pumariega}, Wayne Chen, Alexander~M. Rush,
  Wenting Zhao, and Sanjiban Choudhury.
\newblock Multi-{{Turn Code Generation Through Single-Step Rewards}}, June
  2025.

\bibitem[Jain et~al.(2024{\natexlab{a}})Jain, Han, Gu, Li, Yan, Zhang, Wang,
  {Solar-Lezama}, Sen, and Stoica]{jain2024}
Naman Jain, King Han, Alex Gu, Wen-Ding Li, Fanjia Yan, Tianjun Zhang, Sida
  Wang, Armando {Solar-Lezama}, Koushik Sen, and Ion Stoica.
\newblock {{LiveCodeBench}}: {{Holistic}} and {{Contamination Free Evaluation}}
  of {{Large Language Models}} for {{Code}}.
\newblock Technical Report arXiv:2403.07974, arXiv, June 2024{\natexlab{a}}.
\newblock URL \url{http://arxiv.org/abs/2403.07974}.

\bibitem[Jain et~al.(2024{\natexlab{b}})Jain, Han, Gu, Li, Yan, Zhang, Wang,
  {Solar-Lezama}, Sen, and Stoica]{jain:livecodebench}
Naman Jain, King Han, Alex Gu, Wen-Ding Li, Fanjia Yan, Tianjun Zhang, Sida
  Wang, Armando {Solar-Lezama}, Koushik Sen, and Ion Stoica.
\newblock {{LiveCodeBench}}: {{Holistic}} and {{Contamination Free Evaluation}}
  of {{Large Language Models}} for {{Code}}.
\newblock In \emph{International {{Conference}} on {{Learning
  Representationsm}} ({{ICLR}})}, October 2024{\natexlab{b}}.

\bibitem[{Kimi Team}(2025)]{kimi-k2}
{Kimi Team}.
\newblock Kimi {{K2}}: {{Open Agentic Intelligence}}, 2025.

\bibitem[Kwon et~al.(2023)Kwon, Li, Zhuang, Sheng, Zheng, Yu, Gonzalez, Zhang,
  and Stoica]{kwon2023efficient}
Woosuk Kwon, Zhuohan Li, Siyuan Zhuang, Ying Sheng, Lianmin Zheng, Cody~Hao Yu,
  Joseph~E. Gonzalez, Hao Zhang, and Ion Stoica.
\newblock Efficient memory management for large language model serving with
  pagedattention.
\newblock In \emph{Proceedings of the ACM SIGOPS 29th Symposium on Operating
  Systems Principles}, 2023.

\bibitem[Loshchilov \& Hutter(2019)Loshchilov and Hutter]{loshchilov2019pre}
Ilya Loshchilov and Frank Hutter.
\newblock Decoupled {{Weight Decay Regularization}}, January 2019.
\newblock URL \url{http://arxiv.org/abs/1711.05101}.

\bibitem[Lozhkov et~al.(2024{\natexlab{a}})Lozhkov, Li, Allal, Cassano,
  {Lamy-Poirier}, Tazi, Tang, Pykhtar, Liu, Wei, Liu, Tian, Kocetkov, Zucker,
  Belkada, Wang, Liu, Abulkhanov, Paul, Li, Li, Risdal, Li, Zhu, Zhuo,
  Zheltonozhskii, Dade, Yu, Krau{\ss}, Jain, Su, He, Dey, Abati, Chai,
  Muennighoff, Tang, Oblokulov, Akiki, Marone, Mou, Mishra, Gu, Hui, Dao,
  Zebaze, Dehaene, Patry, Xu, McAuley, Hu, Scholak, Paquet, Robinson, Anderson,
  Chapados, Patwary, Tajbakhsh, Jernite, Ferrandis, Zhang, Hughes, Wolf, Guha,
  {von Werra}, and {de Vries}]{lozhkov2024}
Anton Lozhkov, Raymond Li, Loubna~Ben Allal, Federico Cassano, Joel
  {Lamy-Poirier}, Nouamane Tazi, Ao~Tang, Dmytro Pykhtar, Jiawei Liu, Yuxiang
  Wei, Tianyang Liu, Max Tian, Denis Kocetkov, Arthur Zucker, Younes Belkada,
  Zijian Wang, Qian Liu, Dmitry Abulkhanov, Indraneil Paul, Zhuang Li, Wen-Ding
  Li, Megan Risdal, Jia Li, Jian Zhu, Terry~Yue Zhuo, Evgenii Zheltonozhskii,
  Nii Osae~Osae Dade, Wenhao Yu, Lucas Krau{\ss}, Naman Jain, Yixuan Su, Xuanli
  He, Manan Dey, Edoardo Abati, Yekun Chai, Niklas Muennighoff, Xiangru Tang,
  Muhtasham Oblokulov, Christopher Akiki, Marc Marone, Chenghao Mou, Mayank
  Mishra, Alex Gu, Binyuan Hui, Tri Dao, Armel Zebaze, Olivier Dehaene, Nicolas
  Patry, Canwen Xu, Julian McAuley, Han Hu, Torsten Scholak, Sebastien Paquet,
  Jennifer Robinson, Carolyn~Jane Anderson, Nicolas Chapados, Mostofa Patwary,
  Nima Tajbakhsh, Yacine Jernite, Carlos~Mu{\~n}oz Ferrandis, Lingming Zhang,
  Sean Hughes, Thomas Wolf, Arjun Guha, Leandro {von Werra}, and Harm {de
  Vries}.
\newblock {{StarCoder}} 2 and {{The Stack}} v2: {{The Next Generation}}.
\newblock Technical Report arXiv:2402.19173, arXiv, February
  2024{\natexlab{a}}.
\newblock URL \url{http://arxiv.org/abs/2402.19173}.

\bibitem[Lozhkov et~al.(2024{\natexlab{b}})Lozhkov, Li, Allal, Cassano,
  {Lamy-Poirier}, Tazi, Tang, Pykhtar, Liu, Wei, Liu, Tian, Kocetkov, Zucker,
  Belkada, Wang, Liu, Abulkhanov, Paul, Li, Li, Risdal, Li, Zhu, Zhuo,
  Zheltonozhskii, Dade, Yu, Krau{\ss}, Jain, Su, He, Dey, Abati, Chai,
  Muennighoff, Tang, Oblokulov, Akiki, Marone, Mou, Mishra, Gu, Hui, Dao,
  Zebaze, Dehaene, Patry, Xu, McAuley, Hu, Scholak, Paquet, Robinson, Anderson,
  Chapados, Patwary, Tajbakhsh, Jernite, Ferrandis, Zhang, Hughes, Wolf, Guha,
  {von Werra}, and {de Vries}]{lozhkov:starcoder2}
Anton Lozhkov, Raymond Li, Loubna~Ben Allal, Federico Cassano, Joel
  {Lamy-Poirier}, Nouamane Tazi, Ao~Tang, Dmytro Pykhtar, Jiawei Liu, Yuxiang
  Wei, Tianyang Liu, Max Tian, Denis Kocetkov, Arthur Zucker, Younes Belkada,
  Zijian Wang, Qian Liu, Dmitry Abulkhanov, Indraneil Paul, Zhuang Li, Wen-Ding
  Li, Megan Risdal, Jia Li, Jian Zhu, Terry~Yue Zhuo, Evgenii Zheltonozhskii,
  Nii Osae~Osae Dade, Wenhao Yu, Lucas Krau{\ss}, Naman Jain, Yixuan Su, Xuanli
  He, Manan Dey, Edoardo Abati, Yekun Chai, Niklas Muennighoff, Xiangru Tang,
  Muhtasham Oblokulov, Christopher Akiki, Marc Marone, Chenghao Mou, Mayank
  Mishra, Alex Gu, Binyuan Hui, Tri Dao, Armel Zebaze, Olivier Dehaene, Nicolas
  Patry, Canwen Xu, Julian McAuley, Han Hu, Torsten Scholak, Sebastien Paquet,
  Jennifer Robinson, Carolyn~Jane Anderson, Nicolas Chapados, Mostofa Patwary,
  Nima Tajbakhsh, Yacine Jernite, Carlos~Mu{\~n}oz Ferrandis, Lingming Zhang,
  Sean Hughes, Thomas Wolf, Arjun Guha, Leandro {von Werra}, and Harm {de
  Vries}.
\newblock {{StarCoder}} 2 and {{The Stack}} v2: {{The Next Generation}},
  February 2024{\natexlab{b}}.

\bibitem[Luo et~al.(2023)Luo, Xu, Zhao, Sun, Geng, Hu, Tao, Ma, Lin, and
  Jiang]{ziyang:wizard-coder}
Ziyang Luo, Can Xu, Pu~Zhao, Qingfeng Sun, Xiubo Geng, Wenxiang Hu, Chongyang
  Tao, Jing Ma, Qingwei Lin, and Daxin Jiang.
\newblock {{WizardCoder}}: {{Empowering Code Large Language Models}} with
  {{Evol-Instruct}}, June 2023.

\bibitem[Ma et~al.(2023)Ma, Liu, Yu, Zhang, Jiang, Wang, and
  Li]{yingwei:pretraining-on-code}
Yingwei Ma, Yue Liu, Yue Yu, Yuanliang Zhang, Yu~Jiang, Changjian Wang, and
  Shanshan Li.
\newblock At {{Which Training Stage Does Code Data Help LLMs Reasoning}}?
\newblock In \emph{The {{Twelfth International Conference}} on {{Learning
  Representations}}}, October 2023.

\bibitem[Microsoft(2025)]{microsoft2025phi4minitechnicalreportcompact}
Microsoft.
\newblock Phi-4-mini technical report: Compact yet powerful multimodal language
  models via mixture-of-loras, 2025.
\newblock URL \url{https://arxiv.org/abs/2503.01743}.

\bibitem[Moritz et~al.(2018)Moritz, Nishihara, Wang, Tumanov, Liaw, Liang,
  Elibol, Yang, Paul, Jordan, and Stoica]{moritz:ray}
Philipp Moritz, Robert Nishihara, Stephanie Wang, Alexey Tumanov, Richard Liaw,
  Eric Liang, Melih Elibol, Zongheng Yang, William Paul, Michael~I. Jordan, and
  Ion Stoica.
\newblock Ray: a distributed framework for emerging ai applications.
\newblock In \emph{Proceedings of the 13th USENIX Conference on Operating
  Systems Design and Implementation}, OSDI'18, pp.\  561–577, USA, 2018.
  USENIX Association.
\newblock ISBN 9781931971478.

\bibitem[Nichols et~al.(2024)Nichols, Polasam, Menon, Marathe, Gamblin, and
  Bhatele]{nichols:rlpf}
Daniel Nichols, Pranav Polasam, Harshitha Menon, Aniruddha Marathe, Todd
  Gamblin, and Abhinav Bhatele.
\newblock Performance-{{Aligned LLMs}} for {{Generating Fast Code}}, April
  2024.

\bibitem[{OCI}(2025)]{oci}
{OCI}.
\newblock Open containers initiative.
\newblock \url{https://opencontainers.org/}, 2025.

\bibitem[Orlanski et~al.(2023)Orlanski, Xiao, Garcia, Hui, Howland, Malmaud,
  Austin, Singh, and Catasta]{orlanski2023pre}
Gabriel Orlanski, Kefan Xiao, Xavier Garcia, Jeffrey Hui, Joshua Howland,
  Jonathan Malmaud, Jacob Austin, Rishabh Singh, and Michele Catasta.
\newblock Measuring {{The Impact Of Programming Language Distribution}}, May
  2023.
\newblock URL \url{http://arxiv.org/abs/2302.01973}.

\bibitem[Penedo et~al.(2025)Penedo, Lozhkov, Kydlíček, Allal, Beeching,
  Lajarín, Gallouédec, Habib, Tunstall, and von Werra]{penedo2025codeforces}
Guilherme Penedo, Anton Lozhkov, Hynek Kydlíček, Loubna~Ben Allal, Edward
  Beeching, Agustín~Piqueres Lajarín, Quentin Gallouédec, Nathan Habib,
  Lewis Tunstall, and Leandro von Werra.
\newblock Open-r1 codeforces.
\newblock \url{https://huggingface.co/datasets/open-r1/codeforces}, 2025.

\bibitem[Roziere et~al.(2021)Roziere, Zhang, Charton, Harman, Synnaeve, and
  Lample]{roziere:transcoder-st}
Baptiste Roziere, Jie Zhang, Francois Charton, Mark Harman, Gabriel Synnaeve,
  and Guillaume Lample.
\newblock Leveraging {{Automated Unit Tests}} for {{Unsupervised Code
  Translation}}.
\newblock In \emph{International {{Conference}} on {{Learning Representations}}
  ({{ICLR}})}, October 2021.

\bibitem[Shao et~al.(2024)Shao, Wang, Zhu, Xu, Song, Bi, Zhang, Zhang, Li, Wu,
  and Guo]{shao2024}
Zhihong Shao, Peiyi Wang, Qihao Zhu, Runxin Xu, Junxiao Song, Xiao Bi, Haowei
  Zhang, Mingchuan Zhang, Y.~K. Li, Y.~Wu, and Daya Guo.
\newblock {{DeepSeekMath}}: {{Pushing}} the {{Limits}} of {{Mathematical
  Reasoning}} in {{Open Language Models}}.
\newblock Technical Report arXiv:2402.03300, arXiv, April 2024.
\newblock URL \url{http://arxiv.org/abs/2402.03300}.

\bibitem[{SmolLM3 Team}(2025)]{smollm3}
{SmolLM3 Team}.
\newblock Smollm3.
\newblock \url{https://smollm3.com/}, 2025.

\bibitem[Tange(2023)]{tange2023}
Ole Tange.
\newblock {{GNU Parallel}} 20231122 ('{{Grindav{\'i}k}}').
\newblock Zenodo, November 2023.
\newblock URL \url{https://zenodo.org/records/10199085}.

\bibitem[Wang et~al.(2023)Wang, Kordi, Mishra, Liu, Smith, Khashabi, and
  Hajishirzi]{wang-etal-2023-self-instruct}
Yizhong Wang, Yeganeh Kordi, Swaroop Mishra, Alisa Liu, Noah~A. Smith, Daniel
  Khashabi, and Hannaneh Hajishirzi.
\newblock Self-instruct: Aligning language models with self-generated
  instructions.
\newblock In Anna Rogers, Jordan Boyd-Graber, and Naoaki Okazaki (eds.),
  \emph{Proceedings of the 61st Annual Meeting of the Association for
  Computational Linguistics (Volume 1: Long Papers)}, pp.\  13484--13508,
  Toronto, Canada, July 2023. Association for Computational Linguistics.
\newblock \doi{10.18653/v1/2023.acl-long.754}.
\newblock URL \url{https://aclanthology.org/2023.acl-long.754/}.

\bibitem[Wei et~al.(2024{\natexlab{a}})Wei, Cassano, Liu, Ding, Jain, Mueller,
  de~Vries, von Werra, Guha, and Zhang]{wei2024}
Yuxiang Wei, Federico Cassano, Jiawei Liu, Yifeng Ding, Naman Jain, Zachary
  Mueller, Harm de~Vries, Leandro von Werra, Arjun Guha, and Lingming Zhang.
\newblock {{SelfCodeAlign}}: {{Self-Alignment}} for {{Code Generation}}.
\newblock \emph{CoRR}, abs/2410.24198, 2024{\natexlab{a}}.
\newblock \doi{10.48550/ARXIV.2410.24198}.
\newblock URL \url{https://doi.org/10.48550/arXiv.2410.24198}.

\bibitem[Wei et~al.(2024{\natexlab{b}})Wei, Cassano, Liu, Ding, Jain, Mueller,
  de~Vries, Werra, Guha, and Zhang]{wei:starcoder2-self-instruct}
Yuxiang Wei, Federico Cassano, Jiawei Liu, Yifeng Ding, Naman Jain, Zachary
  Mueller, Harm de~Vries, Leandro~Von Werra, Arjun Guha, and Lingming Zhang.
\newblock Fully {{Transparent Self-Alignment}} for {{Code Generation}}.
\newblock In \emph{Neural {{Information Processing Systems}} ({{NeurIPS}})},
  2024{\natexlab{b}}.

\bibitem[Wei et~al.(2024{\natexlab{c}})Wei, Wang, Liu, Ding, and
  Zhang]{wei:magicoder}
Yuxiang Wei, Zhe Wang, Jiawei Liu, Yifeng Ding, and Lingming Zhang.
\newblock Magicoder: {{Empowering Code Generation}} with {{OSS-Instruct}}.
\newblock In \emph{International {{Conference}} on {{Machine Learning}}
  ({{ICML}})}, June 2024{\natexlab{c}}.

\bibitem[Wei et~al.(2025)Wei, Duchenne, Copet, Carbonneaux, Zhang, Fried,
  Synnaeve, Singh, and Wang]{wei:swerl}
Yuxiang Wei, Olivier Duchenne, Jade Copet, Quentin Carbonneaux, Lingming Zhang,
  Daniel Fried, Gabriel Synnaeve, Rishabh Singh, and Sida~I. Wang.
\newblock {{SWE-RL}}: {{Advancing LLM Reasoning}} via {{Reinforcement
  Learning}} on {{Open Software Evolution}}, February 2025.

\bibitem[Wolf et~al.(2020)Wolf, Debut, Sanh, Chaumond, Delangue, Moi, Cistac,
  Rault, Louf, Funtowicz, Davison, Shleifer, von Platen, Ma, Jernite, Plu, Xu,
  Scao, Gugger, Drame, Lhoest, and Rush]{wolf-etal-2020-transformers}
Thomas Wolf, Lysandre Debut, Victor Sanh, Julien Chaumond, Clement Delangue,
  Anthony Moi, Pierric Cistac, Tim Rault, Rémi Louf, Morgan Funtowicz, Joe
  Davison, Sam Shleifer, Patrick von Platen, Clara Ma, Yacine Jernite, Julien
  Plu, Canwen Xu, Teven~Le Scao, Sylvain Gugger, Mariama Drame, Quentin Lhoest,
  and Alexander~M. Rush.
\newblock Transformers: State-of-the-art natural language processing.
\newblock In \emph{Proceedings of the 2020 Conference on Empirical Methods in
  Natural Language Processing: System Demonstrations}, pp.\  38--45, Online,
  October 2020. Association for Computational Linguistics.
\newblock URL \url{https://www.aclweb.org/anthology/2020.emnlp-demos.6}.

\bibitem[Xie et~al.(2024)Xie, Naik, Fried, and Rose]{cmtrans}
Yiqing Xie, Atharva Naik, Daniel Fried, and Carolyn Rose.
\newblock Data augmentation for code translation with comparable corpora and
  multiple references.
\newblock In \emph{Findings of EMNLP}, 2024.

\bibitem[Xu et~al.(2022)Xu, Alon, Neubig, and Hellendoorn]{polycoder}
Frank~F. Xu, Uri Alon, Graham Neubig, and Vincent~J. Hellendoorn.
\newblock A {{Systematic Evaluation}} of {{Large Language Models}} of {{Code}}.
\newblock In \emph{Deep {{Learning}} for {{Code Workshop}} ({{DL4C}})}, 2022.

\bibitem[Yang et~al.(2025)Yang, Li, Yang, Zhang, Hui, Zheng, Yu, Gao, Huang,
  Lv, Zheng, Liu, Zhou, Huang, Hu, Ge, Wei, Lin, Tang, Yang, Tu, Zhang, Yang,
  Yang, Zhou, Zhou, Lin, Dang, Bao, Yang, Yu, Deng, Li, Xue, Li, Zhang, Wang,
  Zhu, Men, Gao, Liu, Luo, Li, Tang, Yin, Ren, Wang, Zhang, Ren, Fan, Su,
  Zhang, Zhang, Wan, Liu, Wang, Cui, Zhang, Zhou, and Qiu]{yang:qwen3}
An~Yang, Anfeng Li, Baosong Yang, Beichen Zhang, Binyuan Hui, Bo~Zheng, Bowen
  Yu, Chang Gao, Chengen Huang, Chenxu Lv, Chujie Zheng, Dayiheng Liu, Fan
  Zhou, Fei Huang, Feng Hu, Hao Ge, Haoran Wei, Huan Lin, Jialong Tang, Jian
  Yang, Jianhong Tu, Jianwei Zhang, Jianxin Yang, Jiaxi Yang, Jing Zhou,
  Jingren Zhou, Junyang Lin, Kai Dang, Keqin Bao, Kexin Yang, Le~Yu, Lianghao
  Deng, Mei Li, Mingfeng Xue, Mingze Li, Pei Zhang, Peng Wang, Qin Zhu, Rui
  Men, Ruize Gao, Shixuan Liu, Shuang Luo, Tianhao Li, Tianyi Tang, Wenbiao
  Yin, Xingzhang Ren, Xinyu Wang, Xinyu Zhang, Xuancheng Ren, Yang Fan, Yang
  Su, Yichang Zhang, Yinger Zhang, Yu~Wan, Yuqiong Liu, Zekun Wang, Zeyu Cui,
  Zhenru Zhang, Zhipeng Zhou, and Zihan Qiu.
\newblock Qwen3 {{Technical Report}}, May 2025.

\bibitem[Yu et~al.(2025)Yu, Zhang, Zhu, Yuan, Zuo, Yue, Dai, Fan, Liu, Liu,
  Liu, Lin, Lin, Ma, Sheng, Tong, Zhang, Zhang, Zhang, Zhu, Zhu, Chen, Chen,
  Wang, Yu, Song, Wei, Zhou, Liu, Ma, Zhang, Yan, Qiao, Wu, and
  Wang]{yu_dapo2025}
Qiying Yu, Zheng Zhang, Ruofei Zhu, Yufeng Yuan, Xiaochen Zuo, Yu~Yue, Weinan
  Dai, Tiantian Fan, Gaohong Liu, Lingjun Liu, Xin Liu, Haibin Lin, Zhiqi Lin,
  Bole Ma, Guangming Sheng, Yuxuan Tong, Chi Zhang, Mofan Zhang, Wang Zhang,
  Hang Zhu, Jinhua Zhu, Jiaze Chen, Jiangjie Chen, Chengyi Wang, Hongli Yu,
  Yuxuan Song, Xiangpeng Wei, Hao Zhou, Jingjing Liu, Wei-Ying Ma, Ya-Qin
  Zhang, Lin Yan, Mu~Qiao, Yonghui Wu, and Mingxuan Wang.
\newblock {{DAPO}}: {{An Open-Source LLM Reinforcement Learning System}} at
  {{Scale}}.
\newblock Technical Report arXiv:2503.14476, arXiv, May 2025.
\newblock URL \url{http://arxiv.org/abs/2503.14476}.

\bibitem[Zeng et~al.(2025)Zeng, Jiang, Wang, Nie, Chen, and
  Chen]{zeng:acecoder}
Huaye Zeng, Dongfu Jiang, Haozhe Wang, Ping Nie, Xiaotong Chen, and Wenhu Chen.
\newblock {{ACECODER}}: {{Acing Coder RL}} via {{Automated Test-Case
  Synthesis}}, May 2025.

\bibitem[Zhou et~al.(2025)Zhou, Jiang, Tian, Weston, Levine, Sukhbaatar, and
  Li]{zhou:sweet-rl}
Yifei Zhou, Song Jiang, Yuandong Tian, Jason Weston, Sergey Levine, Sainbayar
  Sukhbaatar, and Xian Li.
\newblock {{SWEET-RL}}: {{Training Multi-Turn LLM Agents}} on {{Collaborative
  Reasoning Tasks}}, March 2025.

\end{thebibliography}

\clearpage
\appendix

\begin{figure}
\centering
\includegraphics[width=.5\textwidth]{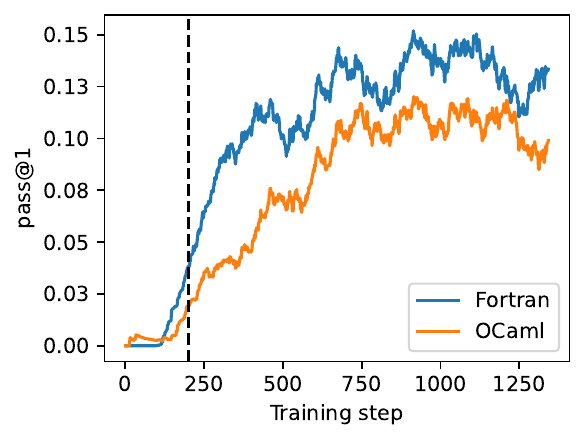}
\caption{Rewards each step from training Qwen 3 4B on Fortran and OCaml. These are very low-resource languages, and rewards are zero for the first several steps.}
\label{fig:ocaml_and_fortran_rewards}
\end{figure}

\section{Preparing Datasets for \OurTechnique}
\label{appendix:reformulation}

\paragraph{\OurDsMbpp}
is a dataset of Mostly Basic Programming Problems,
transformed from the MBPP~\citep{austin2021} dataset.
The source dataset contains problems
asking to complete a Python function from on its signature and a docstring,
where the docstring specifies what result the function should return;
our dataset contains equivalent programs which read/write the same data
from/to standard input/output.
Out of 974 problems, we are able to translate 776 problems into \OurDsMbpp{} (348 problems are in the sanitized subset of MBPP).
We analyzed \OurDsMbpp\ for data similarity against \OurDsLcb\ using \citet{decon} and found no data overlap.
The analysis was done based on 5-grams with no token sampling (see Decon documentation).

Processing the dataset with Qwen3-32B took less than 1 hour using 2 H100 GPUs.
\Cref{fig:mbpp-example,fig:mbpp-reformulated} in the main body of the paper
shows a sample problem from the dataset, before and after the reformulation.
We used the following prompt.

\begin{lstlisting}[style=codeblock]
You are a competitive programming expert.
You are given a problem that asks you to implement a function.
Your task is to translate the description of the problem into a form that accepts one set of function arguments as inputs and return the function return value as output.

Use programming competition style input and outputs -- that is, priorize the use of spaces and newlines to separate inputs and outputs over using commas and parentheses (or other delimiters). Specifically, for 2d lists, you should print them as a list of lists, where the outer lists elements are separated by newlines and the elements of the inner lists are separated by spaces.
For example, a 2d list like [[1, 2], [3, 4]] should be printed as:
1 2
3 4
Do not use any other delimiters.

If there are multiple 2d lists, you should use 2 newlines to separate them.
for example, a 2d list like [[1, 2], [3, 4]] and [[5, 6], [7, 8]] should be printed as:
1 2
3 4

5 6
7 8


If the problem requires outputing decimal numbers, make sure the output format specifies to round all decimal numbers to 4 decimal places. In this case, you should also round all the numbers in the output to 4 decimal places.

Do not forget to specify the input and output format in the description.

Here is the problem description:
{original mbpp problem description}

Here are the test cases:
{original mbpp test cases}

You should return a json object with the following fields:
- "description": the description of the problem
- "input_format": a string describing the input format
- "output_format": a string describing the output format
- "tests": a list of test cases, each test case is a json object with the following fields:
  - "input": a string that represents the input of the test case, in the same format as the input format in the description
  - "output": a string that represents the output of the test case, in the same format as the output format in the description

Place your response in a single ```json ``` block. Do not include any other text in your response.
\end{lstlisting}

\paragraph{\OurDsCodeforces}
is a dataset of competitive programming problems,
created from Codeforces problems in the \texttt{open-r1/codeforces} dataset.
The source problems already specified programs by their I/O behavior,
hence only very minor changes were needed to
build language-universal \OurTechnique\ problems
out of the fields in the dataset:
we only skipped the time and memory restrictions present in the original problems.
To be precise, we used data from the \texttt{open-r1/codeforces-cots} dataset, \texttt{solutions\_py\_decontaminated} subset,
which contains problems decontaminated using 8-gram overlap against multiple benchmarks, in particular LiveCodeBench.
We used the auxilliary \texttt{checker\_interactor} subset
to only keep the problems which admit a simple verifier for their solutions,
i.e., problems where a single output is correct for each input
and where the solution only needs to read data from its input,
compute the result, and write it to the standard output.
We prepared both a train and a test split.
The former contains 5369 problems and the latter
contains 105 problems we held out from the source dataset,
5 selected manually and 100 randomly.
The manually-selected problems were chosen to be much easier than average,
to make it easier to detect if a model can solve any problems at all in a given programming language.
In short, the 5 problems are:
``output a number in binary notation'',
``remove all digits from a string'',
``check if the parentheses are balanced'',
``parse two integers and add them'',
and the following longer problem: ``Petr stands in line of n people, but he doesn't know exactly which position he occupies. He can say that there are no less than a people standing in front of him and no more than b people standing behind him. Find the number of different positions Petr can occupy.''

The train split features randomized prompts,
which we found help with generalizing the results of training on the dataset
to other benchmarks.
The prompts were randomly split into a number of types.
30\% of the prompts use standard Markdown headings to start different sections of the prompt,
35\% use bold text instead,
and the remaining 35\% simply concatenate the prompt sections together.
Most of the prompts follow the source dataset and feature an I/O sample in the prompt,
with other samples withheld as private.
Half of the prompts of the final type do not feature any I/O sample.

\paragraph{\OurDsLcb}
is also a dataset of competitive programming problems,
created from a subset of the LiveCodeBench dataset~\citep{jain2024}.
LiveCodeBench 5.0 has 880 problems, of which 381 have Python starter code and test cases.
The remaining 499 problems do not use starter code and instead use standard I/O to specify and test solutions.
Hence we used these problems to transform LiveCodeBench into an \OurTechnique\ dataset.
\OurDsLcb\ only has a test split, like its source dataset.

\section{\OurTechnique{} Configurations}
\label{appendix:configurations}

In this section, we list the configurations that we use for our target languages.
The configuration files use YAML.
The prompts for OCaml and Fortran have instructions generated by OpenAI o3.

The Lua configuration:
\vskip 1em
\begin{lstlisting}[style=codeblock]
prompt: Use Lua 5.1, targeting LuaJIT.
install: apt-get install -y luajit
filename: snippet.lua
execute: luajit snippet.lua
\end{lstlisting}

\vskip 1em
The Julia configuration:
\vskip 1em
\begin{lstlisting}[style=codeblock]
prompt: Use Julia 1.11.
container:
  base-image: "julia:1.11.3"
  type: debian
filename: snippet.jl
execute: julia snippet.jl
\end{lstlisting}

\vskip 1em
The R configuration (unmodified, unlike \cref{fig:r-config}):
\vskip 1em
\begin{lstlisting}[style=codeblock]
install: apt-get install -y r-cran-tidyverse
filename: snippet.R
execute: Rscript snippet.R
prompt: |
  Use R version 4. Use `readLines(con = file("stdin"))` to read input from stdin. Optionally, use the `n` argument to read the first `n` lines. For example:
  ```r
  input <- readLines(con = file("stdin"), n = 1)
  n <- as.integer(input)
  cat(n) # print the first line of input
  ```
  Also, use `cat` to print output to stdout. For example:
  ```r
  cat(n)
  ```
  Please do not use `print` to print output.
\end{lstlisting}

\vskip 1em
The OCaml configuration:
\vskip 1em
\begin{lstlisting}[style=codeblock]
prompt: |
  Use OCaml 5.

  Numbers:   + - * / mod   vs.   +. -. *. /. **    (add dots!)
  Casts:     float_of_int   int_of_float   int_of_string
  Mutation:  refs (:= !) or pass new values recursively
  Strings:   split_on_char, String.get => char, use Printf "%
  Lists:     avoid List.nth; prefer pattern-match / folds / arrays
container:
  base-image: "docker.io/ocaml/opam:ubuntu-22.04-ocaml-5.0"
  type: debian
install:
  container-instructions: |
    RUN opam install base stdio utop
    ENV OPAM_SWITCH_PREFIX='/home/opam/.opam/5.0'
    ENV CAML_LD_LIBRARY_PATH='/home/opam/.opam/5.0/lib/stublibs:/home/opam/.opam/5.0/lib/ocaml/stublibs:/home/opam/.opam/5.0/lib/ocaml'
    ENV OCAML_TOPLEVEL_PATH='/home/opam/.opam/5.0/lib/toplevel'
    ENV MANPATH=':/home/opam/.opam/5.0/man'
    ENV PATH='/home/opam/.opam/5.0/bin:/usr/local/sbin:/usr/local/bin:/usr/sbin:/usr/bin:/sbin:/bin'
filename: snippet.ml
execute: utop -require base -require stdio snippet.ml
\end{lstlisting}

\vskip 1em
The Fortran configuration:
\vskip 1em
\begin{lstlisting}[style=codeblock]
prompt: |
  Use Fortran 90. Some tips:

  Always begin each scope with implicit none, pick explicit kinds via selected_*_kind, and declare proper lengths-character(len=*) is legal only for dummy arguments, not locals.  Strings are blank-padded: call len_trim before iterating, and store dynamic text in deferred-length allocatables (character(len=:), allocatable :: s).  List-directed read(*,*) arr does not auto-size arrays; read a count first, then allocate and read, or tokenize a line manually.  When translating 0-based formulas (heaps, bit positions) remember Fortran arrays default to 1-based; if you want 0-based, declare lower bounds.  Use real literals (2.0d0, 1.0_rk) to avoid silent integer division, and guard against overflow when exponentiating integers.  For frequency tables, allocate an array or use findloc; Fortran lacks native dicts/sets, so you must implement search yourself.  Prefer array intrinsics (sum, count, pack) over hand-rolled loops, and keep helper procedures inside a contains section or module so interfaces are explicit.  return inside the main program is non-idiomatic; use structured blocks or stop.  Never print interactive prompts in batch solutions; just read, compute, and write.
install: apt-get install -y gfortran
filename: snippet.f90
compile: gfortran -o snippet.out snippet.f90
execute: ./snippet.out
\end{lstlisting}

\section{Training and Results}
\subsection{Choosing hyperparameters}
\label{appendix:choosing-hyperparameters}
Before picking the hyperparameters described in \cref{sec:hyperparameters},
we investigated other values by training the Qwen 3 4B model on a previous version of \OurDsCodeforces.
We trained two models for each of Lua, Julia and R,
using a linear learning rate schedule with the same learning rate.
We decided against it since some of the runs degraded the model's capabilities,
unlike any of the runs we did with a cosine decay schedule.

We compared between GRPO group sizes of 16, 32 and 64
by training the same model on Lua.
In some runs with group size 16, we saw the model improved significantly less than at higher group sizes.
We ran an experiment to compared different temperature and group size settings
(\cref{tab:hyperparam-sweep}).

\begin{table}
\caption{
  Hyperparameter sweep, pass@1 score.
  \label{tab:hyperparam-sweep}
}
\small
\centering
\begin{booktabs}{colspec={@{}l|[gray8]rr|[gray8]r@{}}, colsep=3pt}
  \toprule
  Model & Group size & Temperature & \OurDsLcb\ \\
  \midrule
  Qwen3-4B-CF-Lua & 32 & 0.7 & 23.00 \\
  normal-r1 & 32 & 0.7 & 19.87 \\
  normal-r2 & 32 & 0.7 & 21.58 \\
  size16-r1 & 16 & 0.7 & 19.80 \\
  size16-r2 & 16 & 0.7 & 19.40 \\
  size16-r3 & 16 & 0.7 & 19.65 \\
  size16-r4 & 16 & 0.7 & 20.61 \\
  size64-r1 & 64 & 0.7 & 21.71 \\
  size64-r2 & 64 & 0.7 & 21.21 \\
  size64-r3 & 64 & 0.7 & 20.87 \\
  temp0p2-r1 & 32 & 0.2 & 19.91 \\
  temp0p2-r2 & 32 & 0.2 & 20.39 \\
  temp0p2-r3 & 32 & 0.2 & 21.98 \\
  temp1-r1 & 32 & 1.0 & 21.22 \\
  temp1-r2 & 32 & 1.0 & 21.48 \\
  temp1-r3 & 32 & 1.0 & 20.30 \\
  \bottomrule
\end{booktabs}
\end{table}

The models trained at group size 16 had slightly lower scores compared to other models,
while the ones trained at group size 64 displayed scores comparable to other models.
However, they took significantly longer to train.
Two group size 64 models took $\sim20.5$h to train on average (the third one was trained on a different machine).
In comparison, the group size 32 models trained at the same time on the same machine took $\sim12$h on average.
The models trained at temperatures other than the $0.7$ recommended by the Qwen team
performed similarly to the other models.

As we found no significant difference between the temperature settings and between group sizes 32 and 64,
we chose the smaller group size due to limited resources,
and used the recommended temperature settings.

\subsection{Training dynamics}
\label{appendix:training-curves}

\begin{figure}
\centering
\includegraphics[width=.5\textwidth]{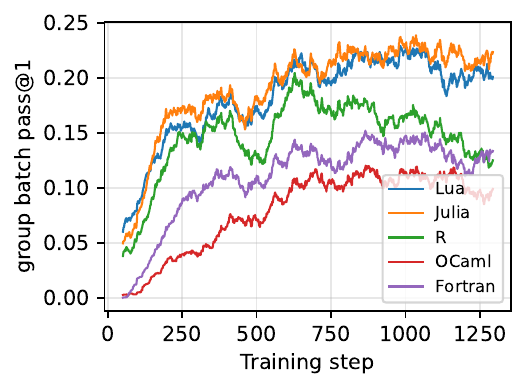}
\caption{Training Qwen3-4B-CF-X, GRPO group batch pass@1.}
\label{fig:curves-all-train}
\end{figure}

\begin{figure}
\centering
\includegraphics[width=.5\textwidth]{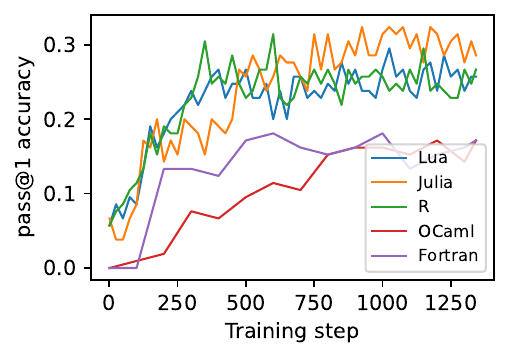}
\caption{Training Qwen3-4B-CF-X, test split pass@1.}
\label{fig:curves-all-test}
\end{figure}

\begin{figure}
\centering
\includegraphics[width=.5\textwidth]{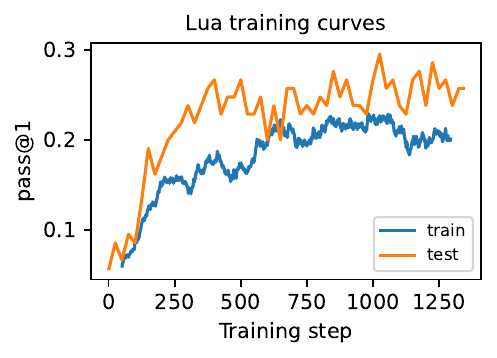}
\caption{Training Qwen3-4B-CF-Lua, GRPO group batch and test split pass@1.}
\label{fig:curves-lua}
\end{figure}

\begin{figure}
\centering
\includegraphics[width=.5\textwidth]{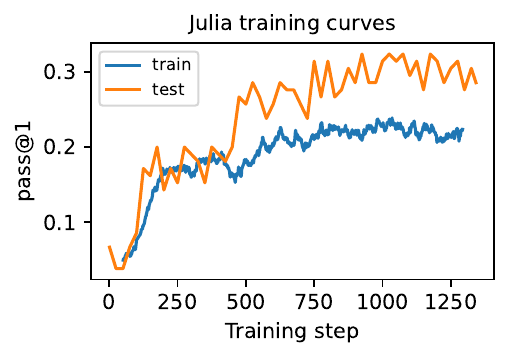}
\caption{Training Qwen3-4B-CF-Julia, GRPO group batch and test split pass@1.}
\label{fig:curves-julia}
\end{figure}

\begin{figure}
\centering
\includegraphics[width=.5\textwidth]{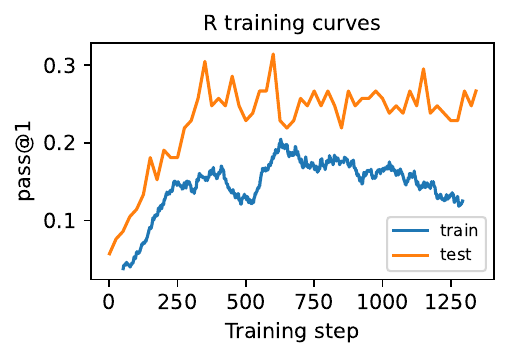}
\caption{Training Qwen3-4B-CF-R, GRPO group batch and test split pass@1.}
\label{fig:curves-r}
\end{figure}

\begin{figure}
\centering
\includegraphics[width=.5\textwidth]{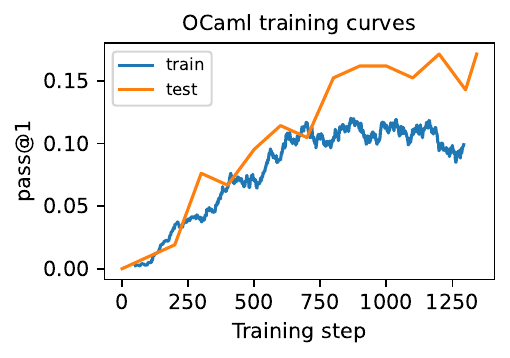}
\caption{Training Qwen3-4B-CF-OCaml, GRPO group batch and test split pass@1.}
\label{fig:curves-ocaml}
\end{figure}

\begin{figure}
\centering
\includegraphics[width=.5\textwidth]{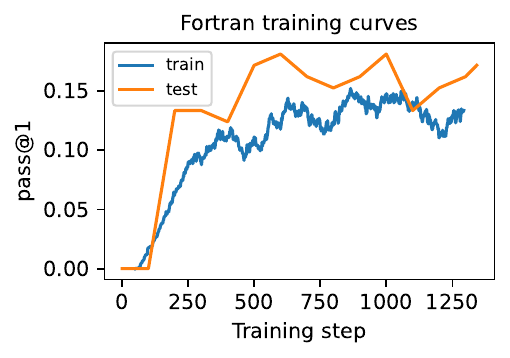}
\caption{Training Qwen3-4B-CF-Fortran, GRPO group batch and test split pass@1.}
\label{fig:curves-fortran}
\end{figure}

In this section we discuss the measurements we took while training the models.
\Cref{fig:curves-all-train} shows the GRPO group batch pass@1 while training the Qwen3-4B-CF-X models.
The scores of all the models are broadly correlated with one another,
which may at least in part be due to training them on the same permutation of the training data.
\Cref{fig:curves-lua,fig:curves-julia,fig:curves-r,fig:curves-ocaml,fig:curves-fortran}
compare the GRPO group pass@1 scores with the pass@1 scores on the test split.
We see that the scores on the test split is broadly correlated with the train split rewards.
In most cases, we see that the train scores keep increasing until the end of the epoch,
together with the test split pass@1 scores,
indicating that the model keeps improving until the end of the dataset.

\subsection{Reward function}
\label{appendix:reward-fn}
We investigated the results of partial rewards.
We trained Qwen 3 4B---the base model of Qwen3-4B-CF-X---on \OurDsCodeforcesS{Lua},
giving it a partial reward of $0.2$ if it generated code which failed one of the tests by producing wrong output but otherwise terminated without an error.
The full reward for a snippet passing all tests was still 1.
\Cref{tab:partial-reward}
shows that the trained models score below Qwen3-4B-CF-Fortran
both on \OurDsLcbS{Lua} and on the test split of \OurDsCodeforcesS{Lua} (counting only the full-credit reward).
The latter scores are particularly far lower, clearly showing that the models learned to abuse the partial-credit reward.

During training, we saw the models focus on the partial reward.
The average result from the partial reward component was clearly increasing more quickly than the result from the full reward component.
In the training generations we inspected, the models also often claimed to generate a ``draft'' answer and
produced a program which ignored the problem in the prompt,
for instance by only printing a hard-coded string such as ``0''.

\begin{table}
\caption{
  Partial reward experiment, pass@1 rates.
  \label{tab:partial-reward}
}
\small
\centering
\begin{booktabs}{colspec={@{}l|[gray8]rr@{}}, colsep=3pt}
  \toprule
  Model & \OurDsLcb\ & \OurDsCodeforces\ (test split) \\
  \midrule
  Qwen3-4B-CF-Lua & 23.00 & 24.76 \\
  partial-r1 & 18.57 & 13.33 \\
  partial-r2 & 20.16 & 15.62 \\
  \bottomrule
\end{booktabs}
\end{table}

\subsection{Cross-Programming-Language Negative Transfer}
\label{appendix:crosspl}
To demonstrate that \OurTechnique\ training does not lower performance on different programming languages,
we evaluated the models we trained on variants of \OurDsLcb\ (\cref{tab:crosspl}).

\begin{table}
\caption{
  Cross-PL evaluation, pass@1 rates.
  \label{tab:crosspl}
}
\small
\centering
\begin{booktabs}{colspec={@{}l|[gray8]rrrr@{}}, colsep=3pt}
  \toprule
  Model & \SetCell[c=4]{c} \OurDsLcb \\
  \SetCell{r} X= & Python & Lua & Julia & R \\
  \midrule
  Qwen 3 4B & 34.34 & 11.00 & 10.00 & 10.00 \\
  \SetRow{azure9}
  Qwen3-4B-CF-Lua & 32.96 & 23.00 & 6.55 & 3.00 \\
  \SetRow{azure9}
  Qwen3-4B-CF-Julia & 35.10 & 8.43 & 22.00 & 3.90 \\
  \SetRow{azure9}
  Qwen3-4B-CF-R & 31.58 & 9.08 & 7.92 & 15.00 \\
  Qwen 3 8B & 33.51 & 11.00 & 9.00 & 9.00 \\
  \SetRow{azure9}
  Qwen3-8B-CF-Lua & 34.29 & 25.00 & 7.44 & 5.96 \\
  \SetRow{azure9}
  Qwen3-8B-CF-Julia & 33.84 & 8.26 & 25.00 & 7.03 \\
  \SetRow{azure9}
  Qwen3-8B-CF-R & 34.49 & 9.90 & 7.34 & 19.00 \\
  DSC 6.7B Ins & 16.86 & 7.89 & 4.79 & 7.97 \\
  \SetRow{azure9}
  DSC-6.7B-Ins-CF-Lua & 17.63 & 8.93 & 8.60 & 8.08 \\
  \SetRow{azure9}
  DSC-6.7B-Ins-CF-Julia & 17.60 & 9.54 & 9.13 & 8.56 \\
  \SetRow{azure9}
  DSC-6.7B-Ins-CF-R & 17.62 & 8.38 & 5.75 & 9.83 \\
  \seprule
  Phi4 mini ins & 19.87 & 7.95 & 8.10 & 4.69 \\
  \SetRow{azure9}
  Phi4-mini-ins-CF-Lua & 22.16 & 11.80 & 8.02 & 4.56 \\
  \SetRow{azure9}
  Phi4-mini-ins-CF-Julia & 21.15 & 9.10 & 7.69 & 4.22 \\
  \SetRow{azure9}
  Phi4-mini-ins-CF-R & 19.82 & 8.02 & 8.54 & 11.54 \\
  \seprule
  SmolLM3 3B & 20.91 & 1.02 & 2.85 & 2.07 \\
  \SetRow{azure9}
  SmolLM3-3B-CF-Lua & 21.81 & 7.46 & 2.93 & 2.17 \\
  \SetRow{azure9}
  SmolLM3-3B-CF-Julia & 21.58 & 1.53 & 7.83 & 2.39 \\
  \SetRow{azure9}
  SmolLM3-3B-CF-R & 21.63 & 1.30 & 3.30 & 5.90 \\
  \bottomrule
\end{booktabs}
\end{table}

\subsection{Hardware and Software Used}
\label{appendix:used-hardware-software}
We used three machines while working on this paper: B, R1 and R2.
R2 was only used to generate completions of trained models for evaluation,
while B and R1 were used to train models.
Upon publication of the paper, we will publicly release Wandb records
of our training runs, which include the duration and the machine used.

B has
2 Intel Xeon Gold 6342 CPUs @ 2.80GHz,
1008 GB of RAM,
4 NVIDIA H100 80GB,
and uses Ubuntu 22.04.5 LTS.

R1 has
2 AMD EPYC 9454 48-Core CPUs,
8 NVIDIA H100 80GB (with NVLink connections),
2268 GB of RAM,
and uses Ubuntu 22.04.5 LTS.

R2 has 2 Intel(R) Xeon(R) Gold 6326 CPU @ 2.90GHz,
10 NVIDIA RTX A600,
504 GB of RAM,
and uses Ubuntu 22.04.5 LTS.

When developing the \OurTechnique\ framework,
we used the following major Python libraries:
\texttt{ray v2.46.0},
\texttt{torch v2.6.0},
\texttt{transformers v4.54.1},
\texttt{vllm v0.8.5.post1},
\texttt{datasets v3.4.1},
\texttt{wandb 0.19.11}.

\section{Bug Taxonomy}
\label{appendix:app-error-taxonomy}

\subsection{Prompt for Generating Taxonomy}

We used the following instructions to generate the bug taxonomy, followed by a list of faulty R programs.

\begin{quote}
\textbf{Input:} The attached file contains multiple failed R programs (Version 4) with their:
\begin{itemize}
\item Source code
\item  Expected output
\item  Actual standard output
\item  Error messages (where applicable)
\end{itemize}

\textbf{Objective:} Analyze these program failures systematically to create a comprehensive taxonomy of 10-12 bug themes that categorize the underlying causes of failure.

\textbf{Instructions:}

\begin{enumerate}
\item \textbf{Initial Analysis}

    \begin{itemize}
   \item  Read through ALL program examples carefully
   \item  For each failure, identify the root cause (not just the symptom)
   \item  Note any patterns or commonalities across failures
   \end{itemize}

\item \textbf{Taxonomy Development}
\begin{itemize}
   \item  Create 10-12 distinct bug themes that collectively cover all observed failures
   \item  Each theme should represent a fundamental type of programming error or misconception
   \item  Themes should be mutually exclusive when possible, but comprehensive in coverage
   \item  Order themes from most to least frequent (or by logical grouping)
\end{itemize}

\item \textbf{For Each Bug Theme, Provide:}
   \begin{itemize}
       \item Theme Name: A concise, descriptive title
        \item Description: 2-3 sentences explaining the nature of this bug type
    \item Common Symptoms: How these bugs typically manifest (error messages, incorrect output, etc.)
   \item Root Causes: The underlying programming mistakes or misconceptions
   \item Examples: Reference 2-3 specific programs from the file that exhibit this theme
   \item Prevention Tips: Brief advice on how to avoid this type of bug
\end{itemize}

\item \textbf{Constraints:}
\begin{itemize}
   \item  Focus on R-specific issues as well as general programming errors
   \item  Base your taxonomy ONLY on the provided examples
   \item  You may search online ONLY to understand specific R error messages or function behavior, not for existing bug taxonomies
   \item  Ensure every failed program in the file can be classified under at least one theme
\end{itemize}

\item \textbf{Deliverable Format:}
   Present your taxonomy as a numbered list with clear formatting and comprehensive coverage of all observed failure patterns.
   Supply a short explanation for each theme in your taxonomy.

\end{enumerate}
\end{quote}

The prompt produced a taxonomy of 11 bug categories.
We edited these categories and selected 7 categories relevant to us, shown in \S\ref{appendix:app-error-t}.

\subsection{Bug Taxonomy Used For Analysis}
\label{appendix:app-error-t}

The following categories represent the prevalent themes of programming errors we use in our analysis of bugs in model-generated code.
They cover the full spectrum of parse, runtime, and logical failures typically encountered in programming.
The themes are not mutually exclusive; we allow a program to have more than one themes.

\begin{enumerate}
  \item \textbf{Syntax and Typographical Errors}: Missing commas, mismatched parentheses, or other typos that cause compile-time parse errors.
  \item \textbf{Input Reading and Parsing Errors}: Mis-reading or mis-parsing input, leading to empty or malformed variables and subsequent failures.
  \item \textbf{Uninitialized Variables}: References to variables never defined, causing undefined behavior or runtime faults.
  \item \textbf{Data Type and Conversion Errors}: Incorrect casting or type misuse that triggers type errors, warnings, or incorrect results.
  \item \textbf{Function Misuse and Missing Libraries}: Invocations of non‑existent or mis‑parameterized functions, or missing imports/libraries, causing errors.
  \item \textbf{Algorithmic Logic Flaws}: Programs that compile and run but produce wrong answers due to faulty logic or conditions.
  \item \textbf{Output Formatting and Presentation Errors}: Correct computational results, but incorrect due to formatting issues (e.g. missing newlines/spaces or output spec violations).
\end{enumerate}

\subsection{Radar Charts for All Programming Languages}
\begin{figure}[t]
  \centering
  \includegraphics[width=.5\textwidth]{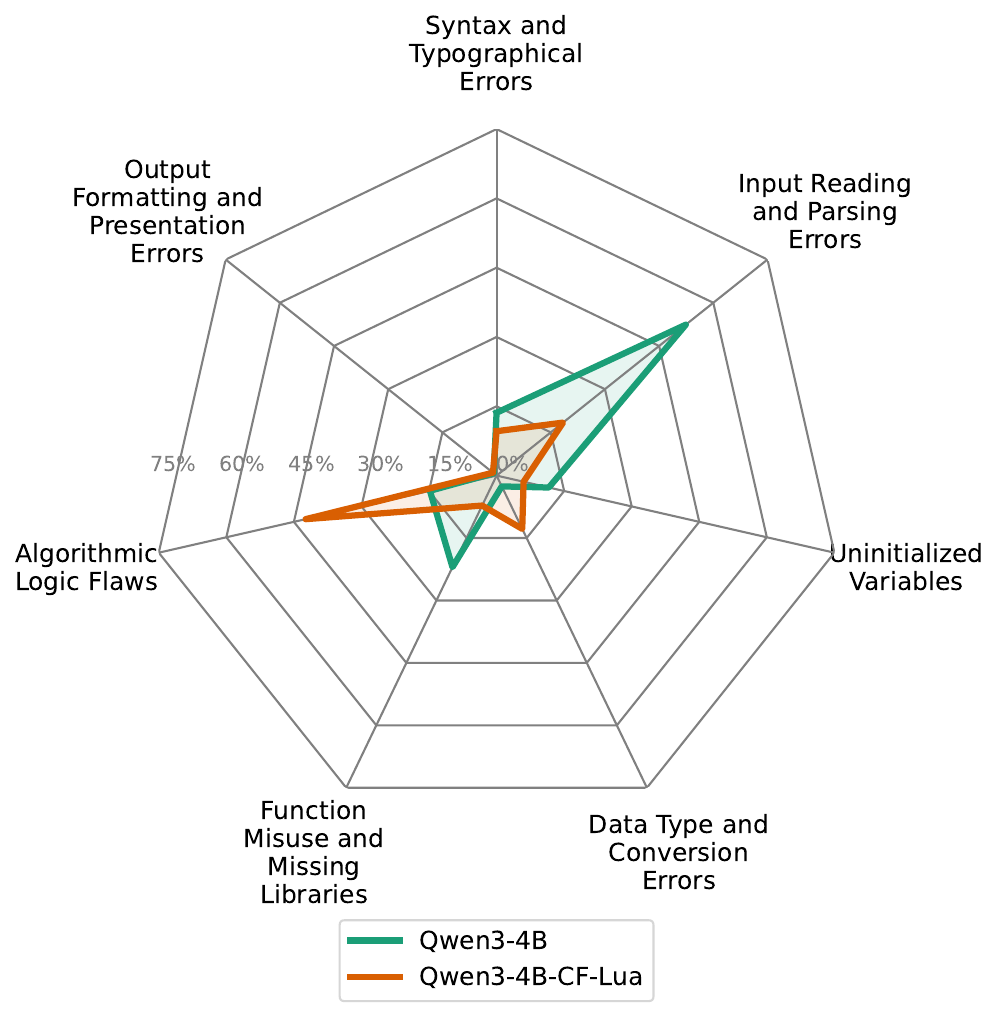}
  \caption{Radar chart of Lua error themes for Qwen3-4B and Qwen3-4B-CF-Lua.}
  \label{fig:app-radar-lua}
\end{figure}

\begin{figure}[t]
  \centering
  \includegraphics[width=.5\textwidth]{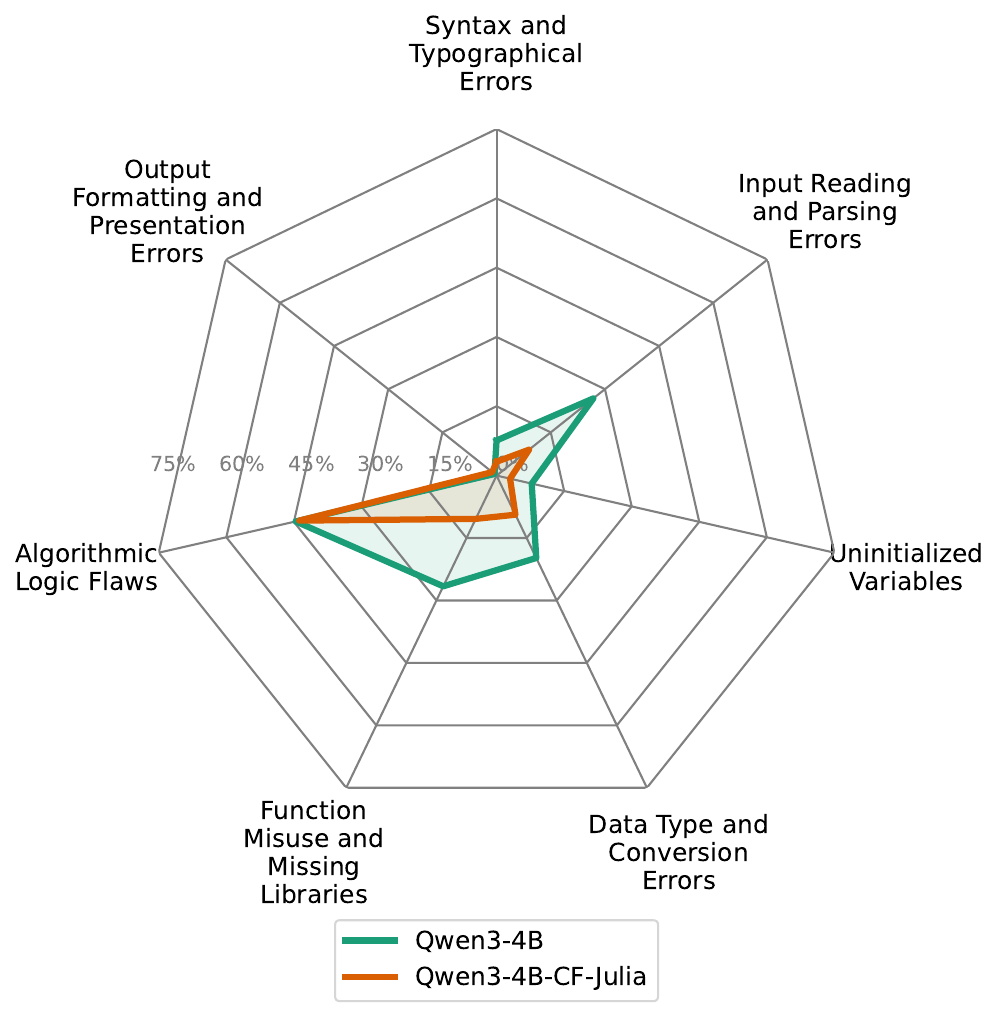}
  \caption{Radar chart of Julia error themes for Qwen3-4B and Qwen3-4B-CF-Julia.}
  \label{fig:app-radar-julia}
\end{figure}

\begin{figure}[t]
  \centering
  \includegraphics[width=.5\textwidth]{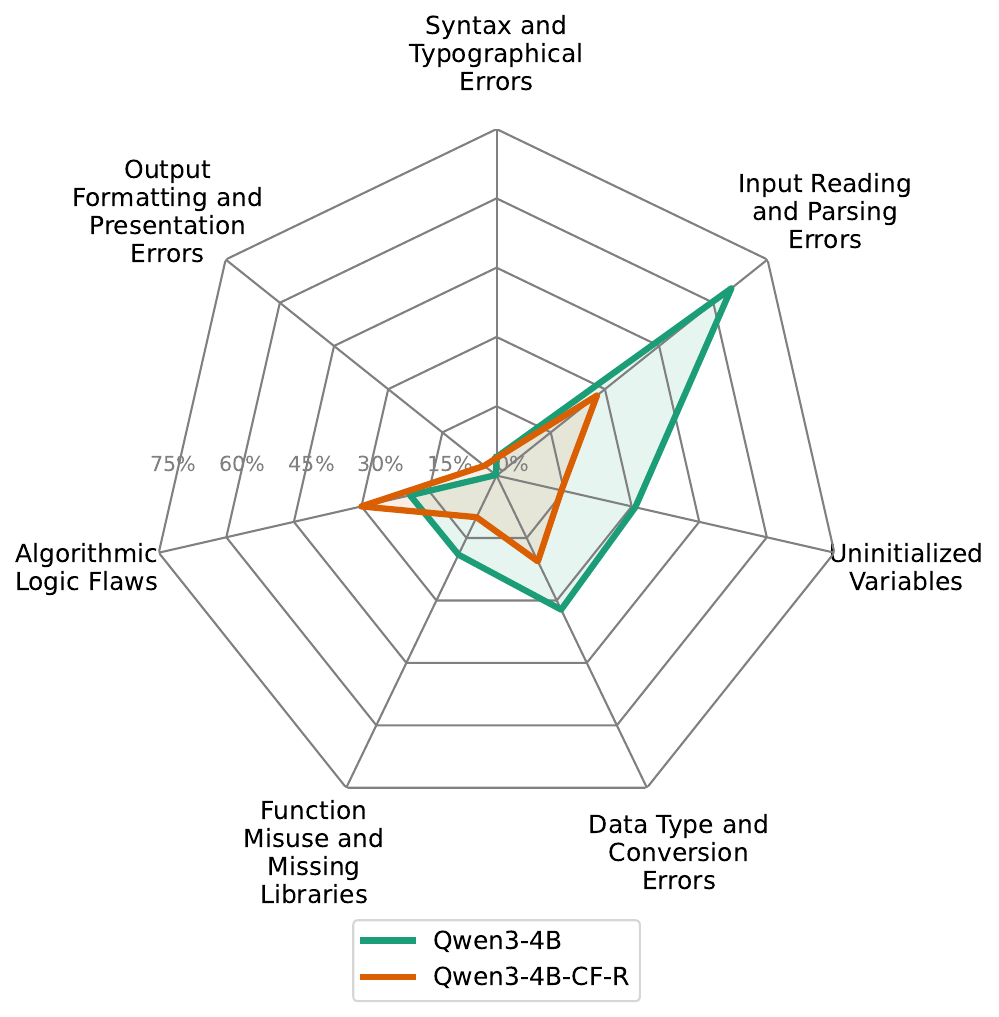}
  \caption{Radar chart of R error themes for Qwen3-4B and Qwen3-4B-CF-R.}
  \label{fig:app-radar-r}
\end{figure}

\begin{figure}[t]
  \centering
  \includegraphics[width=.5\textwidth]{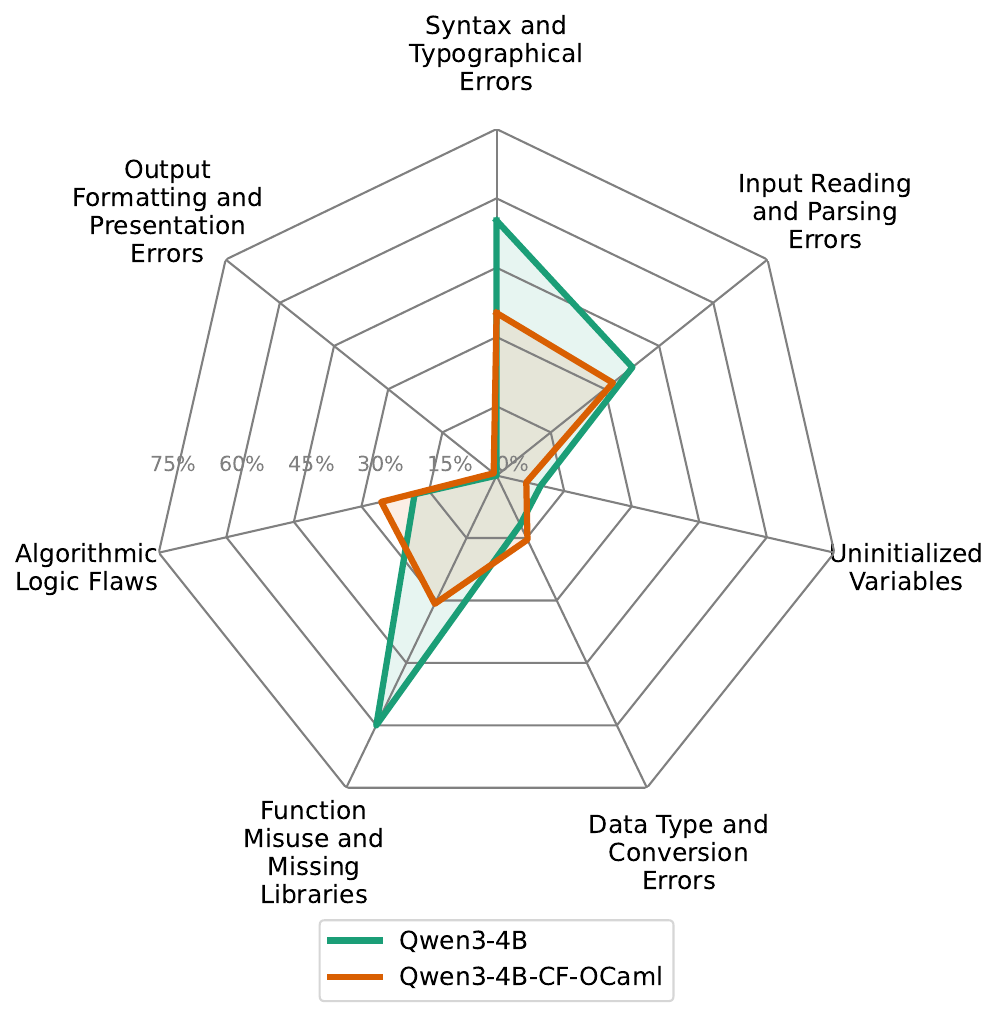}
  \caption{Radar chart of OCaml error themes for Qwen3-4B and Qwen3-4B-CF-OCaml.}
  \label{fig:app-radar-ml}
\end{figure}

\begin{figure}[t]
  \centering
  \includegraphics[width=.5\textwidth]{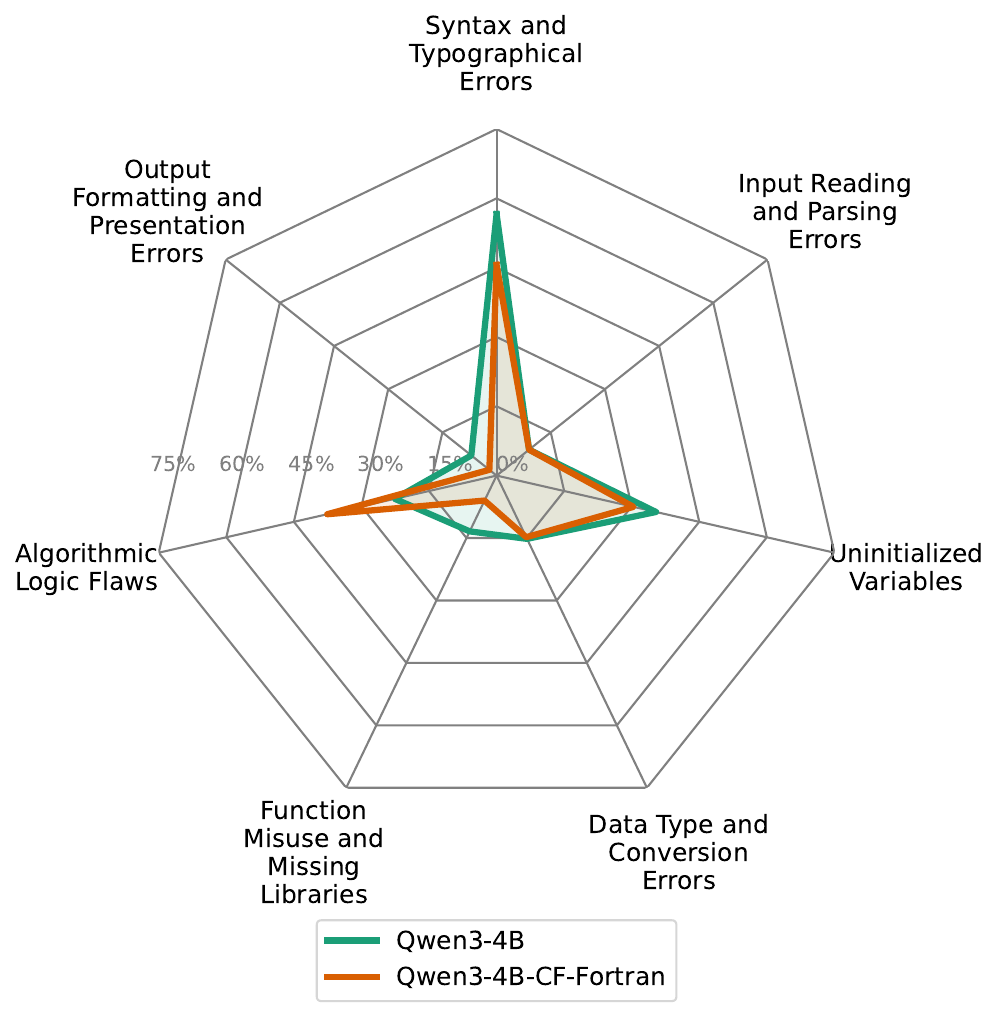}
  \caption{Radar chart of Fortran error themes for Qwen3-4B and Qwen3-4B-CF-Fortran.}
  \label{fig:app-radar-f90}
\end{figure}

Figures~\ref{fig:app-radar-lua}, \ref{fig:app-radar-julia}, \ref{fig:app-radar-r}, \ref{fig:app-radar-ml}, \ref{fig:app-radar-f90} show the error theme charts for all the programming languages we trained a model on. Figure~\ref{fig:app-radar-ml} is the same as Figure~\ref{fig:radar-ml} from the main body; we repeated it here for convenience.

\ifdraft
\section{Aux figures}
This section is draft-only.

\begin{table}
\caption{
  Pass@1 rates on AG-LiveCodeBench-X for Qwen 3 4B models trained on five low-resource languages using \OurDsCodeforces.
  Each model achieves a SOTA result for an open-weight model with ${\le}16$B parameters.
}
\small
\centering
\begin{booktabs}{colspec={@{}l|[gray8]rrr@{}}, colsep=3pt}
  \toprule
  Model & \SetCell[c=3]{c} \OurDsLcb{} \\
  &  X= Lua & Julia & R \\
  \midrule
  Llama-3.3-70B-Ins
              & \textbf{25} & 22 & 13
  \\
  Qwen3-32B
              & 22 & \textbf{26} & 17
  \\
  DSCv2-Lite-Ins-16B
              & 13 & 12 & 9
  \\
  \seprule
  Qwen3-4B
                   & 11 & 10 & 10
  \\
  Qwen3-8B
                   & 11 & 9 & 9
  \\
  \SetRow{azure9}
  Qwen3-4B-CF-X (Ours)
                   & 23 & 22 & 15
  \\
  \SetRow{azure9}
  Qwen3-8B-CF-X (Ours)
                   & \textbf{25} & \textbf{25} & \textbf{19}
  \\
  \bottomrule
\end{booktabs}
\end{table}

\begin{table}
\caption{
  Pass@1 rates on AG-LiveCodeBench-X for Qwen 3 4B models trained on five low-resource languages using \OurDsCodeforces.
  Each model achieves a SOTA result for an open-weight model with ${\le}16$B parameters.
}
\small
\centering
\begin{booktabs}{colspec={@{}l|[gray8]rr@{}}, colsep=3pt}
  \toprule
  Model & \SetCell[c=2]{c} \OurDsLcb{} \\
  &  X= OCaml & Fortran \\
  \midrule
  Llama-3.3-70B-Ins
              & \textbf{7} & 3 \\
  Qwen3-32B
              & 2 & 1 \\
  DSCv2-Lite-Ins-16B
              & \textbf{7} & 6 \\
  \seprule
  Qwen3-4B
              & 1 & 0 \\
  \SetRow{azure9}
  Qwen3-4B-CF-X (Ours)
              & \textbf{7} & \textbf{15} \\
  \bottomrule
\end{booktabs}
\end{table}

\fi

\pagebreak

\end{document}